\documentclass[sigconf]{acmart}

\usepackage{geometry}
\usepackage{graphicx}
\usepackage{float}
\usepackage{caption}
\usepackage{subfig}
\usepackage{bm}
\usepackage{multirow}
\usepackage{makecell}
\usepackage{amsmath}
\usepackage{mathrsfs}
\usepackage[noend]{algpseudocode}
\usepackage{graphicx}
\usepackage{url}
\usepackage[ruled,vlined]{algorithm2e}
\usepackage{color, xcolor}
\newcommand{\code}[1]{{\color{black}\texttt{#1}}} 

\usepackage[justification=justified]{caption}
\renewcommand{\arraystretch}{0.9}

\AtBeginDocument{%
  \providecommand\BibTeX{{%
    \normalfont B\kern-0.5em{\scshape i\kern-0.25em b}\kern-0.8em\TeX}}}

\copyrightyear{2023} 
\acmYear{2023} 
\setcopyright{rightsretained} 
\acmConference[CIKM '23]{Proceedings of the 32nd ACM International Conference on Information and Knowledge Management}{October 21--25, 2023}{Birmingham, United Kingdom} 
\acmBooktitle{Proceedings of the 32nd ACM International Conference on Information and Knowledge Management (CIKM '23), October 21--25, 2023, Birmingham, United Kingdom}
\acmDOI{10.1145/3583780.3614796} 
\acmISBN{979-8-4007-0124-5/23/10}

\usepackage{etoolbox}
\makeatletter
\patchcmd{\maketitle}{\@copyrightpermission}{
   \begin{minipage}{0.3\columnwidth}
     \href{https://creativecommons.org/licenses/by/4.0/}{\includegraphics[width=0.90\textwidth]{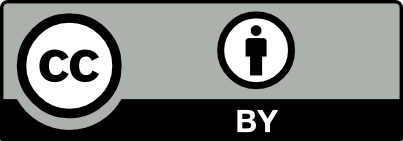}}
   \end{minipage}\hfill
   \begin{minipage}{0.7\columnwidth}
     \href{https://creativecommons.org/licenses/by/4.0/}{This work is licensed under a Creative Commons Attribution International 4.0 License.}
   \end{minipage}
  
}{}{}

\begin{document}
\title[Bridged-GNN: Knowledge Bridge Learning for Effective Knowledge Transfer]{Bridged-GNN: Knowledge Bridge Learning for \\ 
Effective Knowledge Transfer}

\author{Wendong Bi}
\affiliation{%
    \institution{Institute of Computing Technology, University of Chinese Academy of Sciences}
    \city{Beijing}
    \country{China}}
\email{biwendong20g@ict.ac.cn}

\author{Xueqi Cheng}
\authornotemark[1]
\affiliation{%
    \institution{Institute of Computing Technology, Chinese Academy of Sciences}
    \city{Beijing}
    \country{China}}
\email{cxq@ict.ac.cn}

\author{Bingbing Xu}
\authornote{Corresponding authors}
\affiliation{%
    \institution{Institute of Computing Technology, Chinese Academy of Sciences}
    \city{Beijing}
    \country{China}}
\email{xubingbing@ict.ac.cn}

\author{Xiaoqian Sun}
\affiliation{%
    \institution{Institute of Computing Technology, Chinese Academy of Sciences}
    \city{Beijing}
    \country{China}}
\email{sunxiaoqian@ict.ac.cn}

\author{Li Xu}
\affiliation{%
    \institution{Institute of Computing Technology, Chinese Academy of Sciences}
    \city{Beijing}
    \country{China}}
\email{lixu@ict.ac.cn}


\author{Huawei Shen}
\affiliation{%
    \institution{Institute of Computing Technology, Chinese Academy of Sciences}
    \city{Beijing}
    \country{China}}
\email{shenhuawei@ict.ac.cn}





\begin{abstract}
    

Data-hungry problem, characterized by insufficiency and low-quality of data, poses obstacles for deep learning models.
Transfer learning has been a feasible way to transfer knowledge from high-quality external data of source domains to limited data of target domains, 
which follows a domain-level knowledge transfer to learn a shared posterior distribution. However, they are usually built on strong assumptions, e.g., the domain invariant posterior distribution, which is usually unsatisfied and may introduce noises, resulting in poor generalization ability on target domains.
Inspired by Graph Neural Networks (GNNs) that aggregate information from neighboring nodes, we redefine the paradigm as learning a knowledge-enhanced posterior distribution for target domains, namely Knowledge Bridge Learning (KBL). KBL first learns the scope of knowledge transfer by constructing a Bridged-Graph that connects knowledgeable samples to each target sample and then performs sample-wise knowledge transfer via GNNs.
KBL is free from strong assumptions and is robust to noises in the source  data.
Guided by KBL, we propose the \textbf{Bridged-GNN}, including an Adaptive Knowledge Retrieval module to build Bridged-Graph and a Graph Knowledge Transfer module. Comprehensive experiments on both un-relational and relational data-hungry scenarios demonstrate the significant improvements of Bridged-GNN compared with SOTA methods.

\end{abstract}





\maketitle

\section{Introduction}

\begin{figure}[h]
    \centering
    \includegraphics[width=0.94\linewidth]{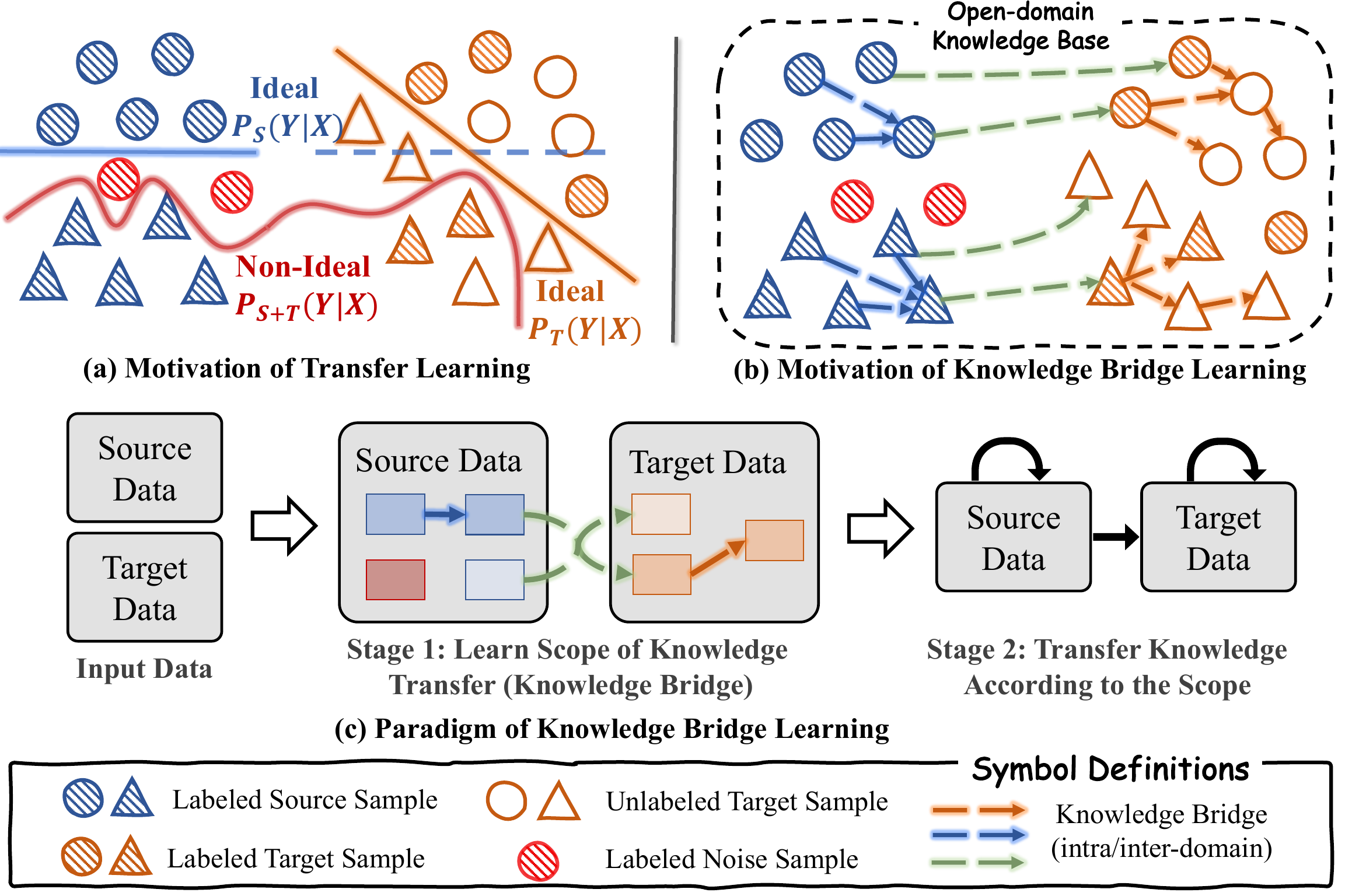}
    \caption{Comparison of Knowledge Bridge Learning (KBL) and  Transfer Learning (TL).}
    \label{fig:intro}
\end{figure}


The data-hungry problem, characterized by insufficiency and low-quality of data, is a prevalent challenge in real-world scenarios. This challenge presents obstacles for deep learning, because they heavily rely on substantial volumes of high-quality data for effective parameter optimization.
Due to the high cost of manual annotation, a feasible way is to find existing high-quality external datasets from other domains to supplement the shortage of target datasets\cite{MME, APE, DANN, Sun2011trading, bi2022company, li2023unipoll}. However, external datasets often exhibit significant domain differences compared to the target dataset (e.g., distribution shifts in labels or features). How to improve the performance of deep learning in data-hungry scenarios with the help of external data is a critical research problem.

Transfer learning (TL)  \cite{DAN, survey_da, gui2022good,wang2022deconfounding, wu2022discovering, wu2023energybased} attempts to transfer knowledge from the source domain data to the target domain data and has gained significant success in various scenarios, e.g., image recognition and text mining \cite{jin2020simple, MME, S3D, zhang2019sequence, chen2009extracting}. 
Specifically, TL methods transfer information at the domain level, i.e., they jointly train models on data from both the source and target domain and diminish the domain difference concurrently. However, these methods have two shortcomings shown in Fig. \ref{fig:intro} (a): (1) they are usually built on strong assumptions \cite{DAN, MME}, e.g., assuming that data from different domains have the same conditional distribution 
(detailed analysis in Sec. \ref{sec:motivation}). These assumptions, however, are usually unsatisfied in real scenarios, thus leading to limited choices for available source domain data or even no satisfied source domain data \cite{survey_da}; (2) domain-level information contains both useful and noisy information, and transferring source information indiscriminately may restrict the modeling capacity.
Due to the above two limitations, the paradigm of domain-level knowledge transfer may lead to poor generalization ability on the target domain. 

Inspired by the idea of Graph Neural Networks (GNNs) that learn node representations by transferring information from neighbors \cite{cai2021graph, Sun2014predict, duval2022higher, ren2022known, AKGR, SymCLKGE},  we redefine the paradigm of knowledge transfer and propose the framework of Knowledge Bridge Learning (\textbf{KBL}). 
In Fig. \ref{fig:intro} (b), we regard both the source domain and target domain as an open-domain knowledge base that consists of diverse samples. Each sample serves as a query, 
which is similar to information retrieval models that retrieve relevant items for a query from a collection of open-domain documents. 
We extract valuable knowledge (samples in the open domain) for each query (a target sample) and bridge them. Such bridges form the scope of knowledge transfer, namely Bridged-Graph,
specifically identifying which samples in the open-domain data contain knowledge that is beneficial for the samples in the target domain.
Then we transfer knowledge under the guidance of the learned scope (Bridged-Graph) with GNNs. 
Overall, KBL consists of two steps as illustrated in Fig. \ref{fig:intro} (c): learning the scope of knowledge transfer (i.e., Bridged-Graph) first and then transferring useful knowledge according to the learned scope. 
 

 
 
The paradigm of KBL has the following unique advantages: (1) KBL breaks the limitations of transfer learning that assume a domain-invariant posterior distribution between the source and target domains. KBL aims at learning a \textbf{knowledge-enhanced posterior distribution} for target domain, i.e., learning $P_T\left(Y|X, \mathcal{K}(X)\right)$ where $\mathcal{K}(X)$ is the knowledge of each sample, rather than jointly learning a shared posterior distribution $P_{S+T}(Y|X)$ of source and target domains. (2) KBL can filter the noise of source domains by  fine-grained knowledge transfer scoped by Bridged-Graph.

Under the paradigm of Knowledge Bridge Learning, we propose a novel Bridged-Graph Neural Network (\textbf{Bridged-GNN}) model. 
Bridged-GNN consists of two main components: the Adaptive Knowledge Retrieval (AKR) module and the Graph Knowledge Transfer (GKT) module. 
The Adaptive Knowledge Retrieval (AKR) module aims to retrieve beneficial samples that contain useful knowledge for the given sample from both the source and target domains. 
Then we view these retrieved beneficial samples, which may be from the source domain (inter-domain) or target domain (intra-domain), as knowledge for each target benefited sample and connect such directed "\textbf{beneficial-benefited}" edges to construct the Bridged-Graph. Bridged-Graph is a structure of data that defines the scope of knowledge transfer. Besides, we propose a scalable training algorithm to optimize the parameters of AKR. In the next step, we use a Graph Knowledge Transfer (GKT) module to transfer sample-wise knowledge on the Bridged-Graph. Here we can use any message-passing-based GNNs as the GKT module.  

\begin{figure}[t]
    \centering
    \includegraphics[width=0.95\linewidth]{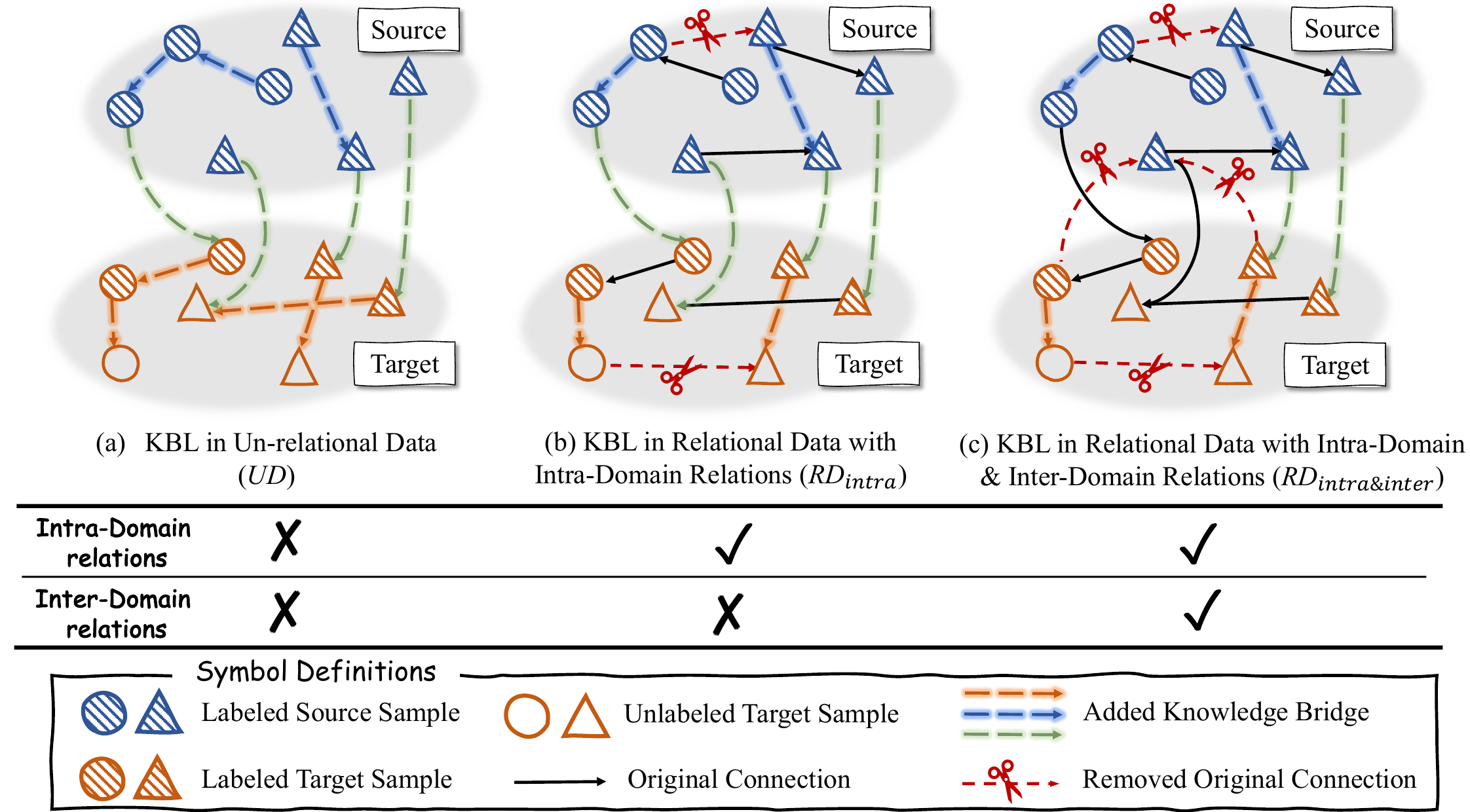}
    \caption{Knowledge Bridge Learning in three scenarios.}
    \label{fig:gkt}
\end{figure}

According to whether there exist intra-domain and inter-domain relations between samples of original data, we divide knowledge transfer into three scenarios (see Fig. \ref{fig:gkt}):  (a) un-relational data, namely $UD$, (b) relational data with intra-domain relations and without inter-domain relations, namely $RD_{intra}$, (c) relational data with both intra-domain relations and inter-domain relations, namely $RD_{intra\&inter}$.
We conduct comprehensive experiments of  three scenarios on four real-world datasets (Company, Twitter, Facebook100, Office31) and a synthetic dataset. The results consistently show that Bridged-GNN gains significant improvements in all three scenarios.


The main contributions of this paper are summarized as follows:
\begin{itemize}
    \item We redefine the paradigm of knowledge transfer as Knowledge Bridge Learning (KBL) that conducts sample-wise knowledge transfers within the learned scope.
    \item We propose a novel Bridged-GNN model under the paradigm of KBL to transfer  knowledge effectively in three different scenarios.
    \item Comprehensive experiments on real-world and synthetic datasets demonstrate the effectiveness of our method.
\end{itemize}
 
\begin{figure*}[h]
    \begin{minipage}[t]{0.18\linewidth}
        \centering
        \subfloat[Twitter]{\includegraphics[width=\linewidth]{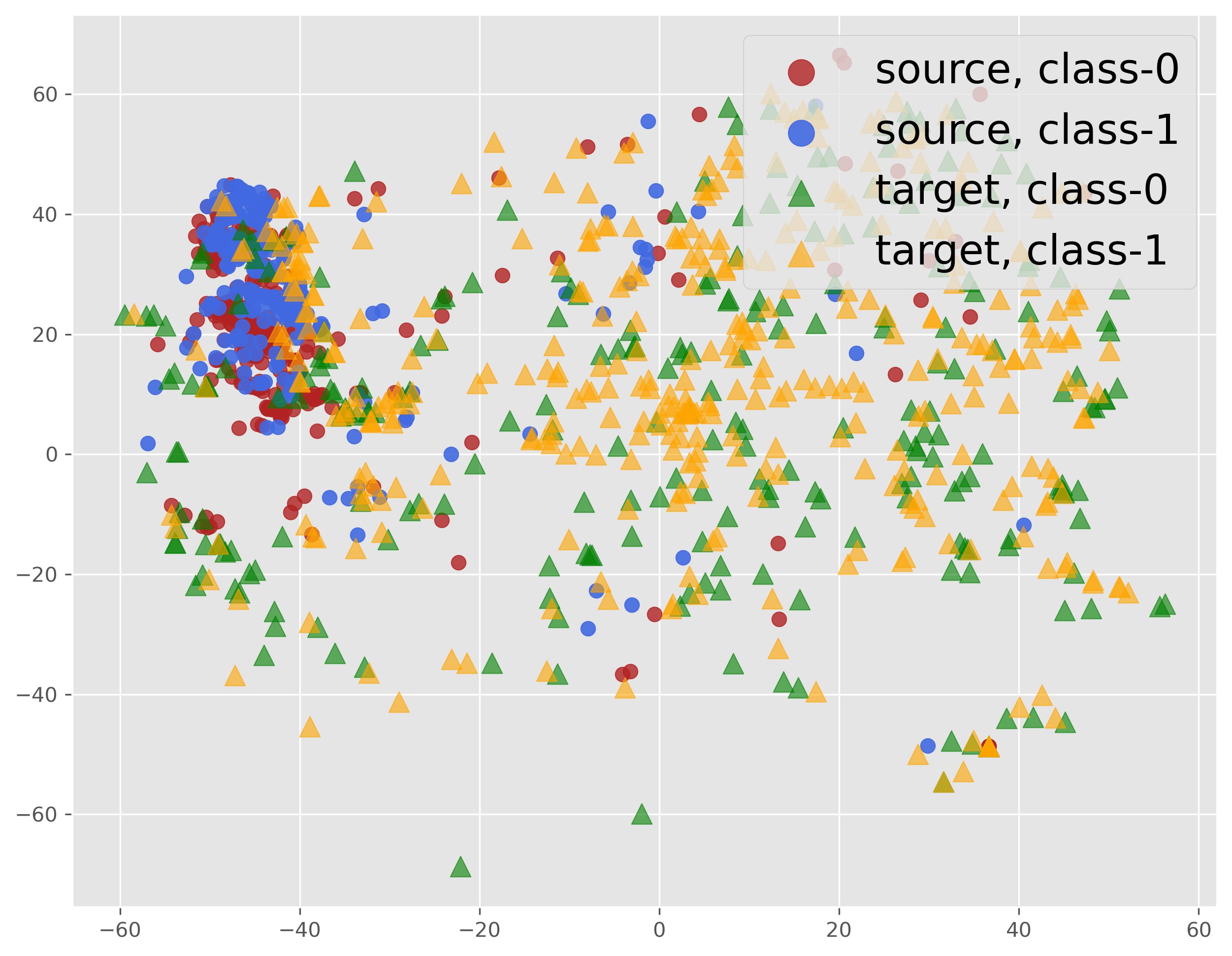}}
    \end{minipage}
    \begin{minipage}[t]{0.18\linewidth}
        \centering
        \subfloat[Company]{\includegraphics[width=\linewidth]{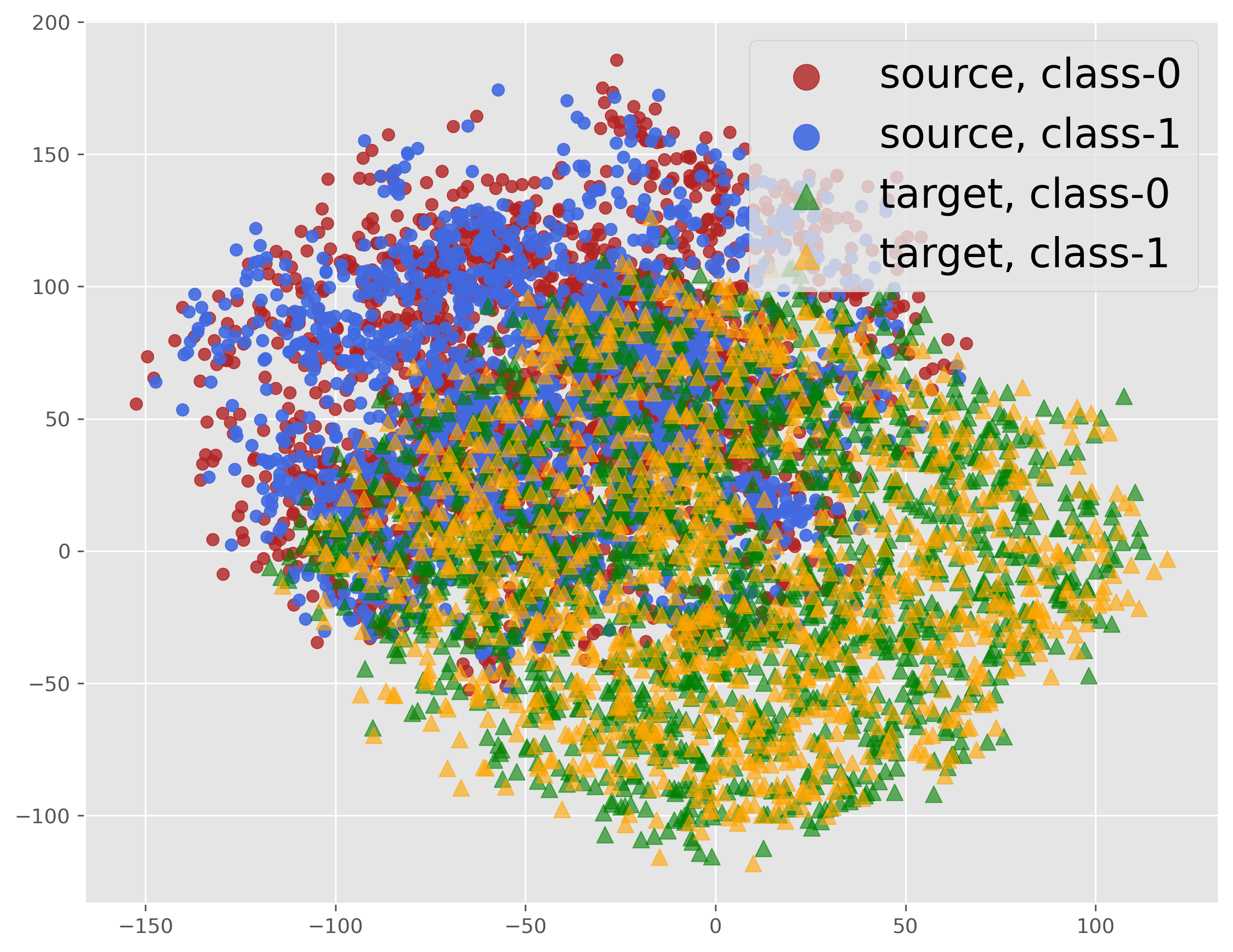}}
    \end{minipage}
    \begin{minipage}[t]{0.18\linewidth}
        \centering
        \subfloat[FB (Hamilton$\rightarrow$Caltech)]{\includegraphics[width=\linewidth]{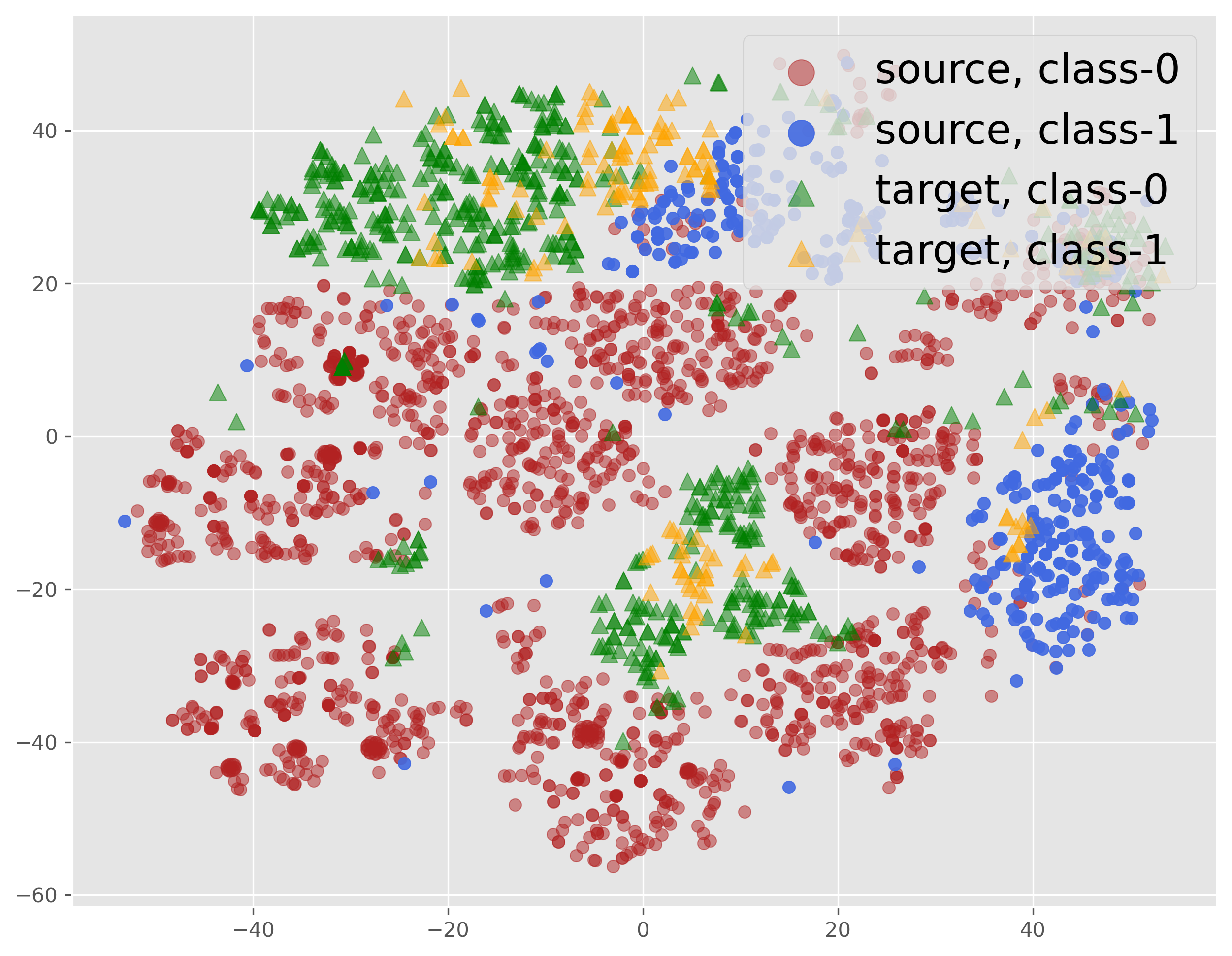}}
    \end{minipage}
    \begin{minipage}[t]{0.18\linewidth}
        \centering
        \subfloat[FB (Howard$\rightarrow$Simmons)]{\includegraphics[width=\linewidth]{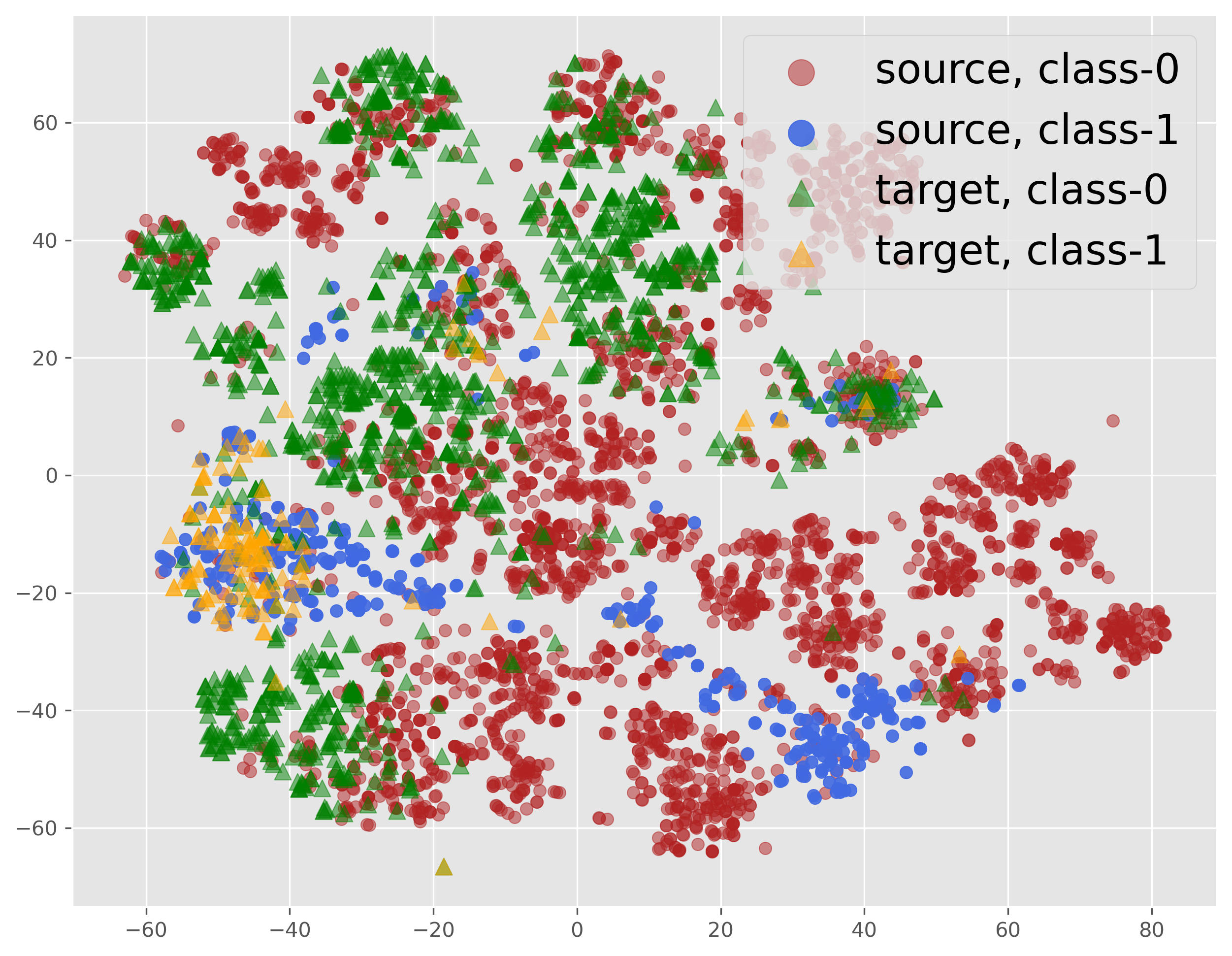}}
    \end{minipage}
    \begin{minipage}[t]{0.18\linewidth}
        \centering
        \subfloat[Sync Dataset]{\includegraphics[width=\linewidth]{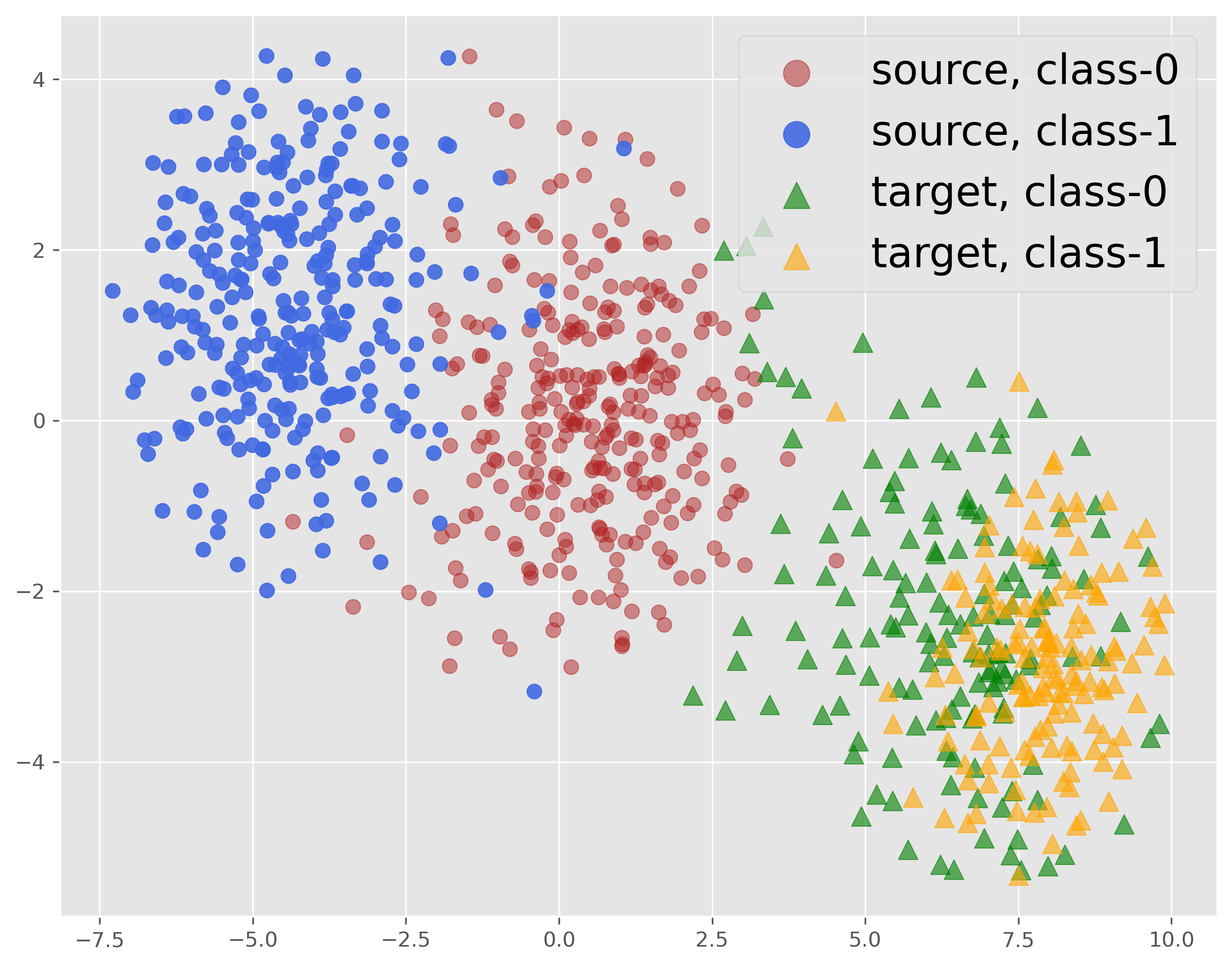}}
    \end{minipage}

    \caption{T-SNE visualization of the domain shift in real-world datasets and synthetic dataset. To better show the difference in feature distribution conditioned on different classes, we only use samples from the first two classes to draw scatter plots.}
    \label{fig:domain_shift_visual}
\end{figure*}

\section{Preliminary}
In this section, we introduce GNNs and our problem definitions.

\subsection{Graph Neural Networks}
Graph, denoted as $G(V, E, X, Y)$, is composed of nodes $V$ and edges $E$, where $V = \{ v_i | v_i\in V \}$ and $E=\{e_{i j}=(v_i, v_j) | v_i \ \text{and}\ \ v_j$ $ \text{is connected} \}$. $N=|V|$ is the number of nodes,  $X\in\mathbb{R}^{N\times D_{in}}$ is the  feature matrix and $Y\in\mathbb{R}^{N\times 1}$ is the labels of all nodes.

Graph Neural Networks (GNNs) aim at learning representations for nodes on the graph. Given a graph $\mathbf{G}(V, E, X, Y)$ as input, GNNs learn node representations by iteratively aggregating information from neighbors. Current mainstream GNNs update node representations with the following function:
\begin{equation}
\label{eq:gnn}
    H^{(l)}_i = \code{Combine}\left( H^{(l-1)}_i, \code{AGG}\big(\{H^{(l-1)}_i, H_j^{(l-1)} | v_j \in \mathcal{N}(v_i) \} \big) \right)
\end{equation}
Where $H^{(l)}_i$ is the node representation vector at $l$-th layer, \code{AGG} denotes the neighborhood aggregation function, and \code{Combine} denotes the combination function that updates node representation with the aggregated neighborhood feature and the central node feature.

\subsection{Problem Definitions}
We introduce the scenarios of Knowledge Transfer and the definitions of Knowledge Bridge Learning in this section.
\subsubsection{Scenarios of Knowledge Transfer}\quad\\
Knowledge transfer can be leveraged in three scenarios as shown in Fig. \ref{fig:gkt}: (a) un-relational data ($UD$) represents the scenario where both source domain and target domain are non-graph data, denoted by $(X^S, Y^S)$ and $(X^T, Y^T)$. (b) data with intra-domain relations and without inter-domain relations ($RD_{intra}$) represents the scenario of both source domain and target data are graph data and there are no edges between the two graphs, denoted by $G^{S}(X^S, Y^S, E^S)$ and $G^T(X^T, Y^T, E^T)$. (c) data with intra-domain relations and inter-domain relations ($RD_{intra\&inter}$) represents the scenario that both intra-domain and inter-domain edges exist in the original graph,
denoted by $G(\{X^S, X^T\}, \{Y^S, Y^T\}, E)$. Existing transfer learning methods usually focus on un-relational scenarios, e.g., images.
\subsubsection{Knowledge Bridge Learning (KBL)}\quad\\
\label{sec:def_kbl}
As shown in Fig. \ref{fig:intro} (c), KBL is a new paradigm of knowledge transfer, which regards both the source domain and target domain as an open-domain knowledge base, and takes each sample as a query.
KBL first learns a scope of knowledge transfer by querying the open-domain knowledge base and then transfers knowledge within the learned scope. In this paper, we propose to use Bridged-Graph to represent the scope of knowledge transfer:
\begin{definition}[Bridged-Graph]
    Bridged-Graph is represented as $G_\text{bridged}=(V^S, V^T, X^S, X^T, Y^S, Y^T, E)$ which defines the scope of knowledge transfer with beneficial-benefited relations between samples, where $V^S$ ($V^T$) is the sample/node set of the source (target) domain, $X^S\in\mathbb{R}^{N^S\times D^{in}}$ ($X^T\in\mathbb{R}^{N^D\times D^{in}}$) is the feature matrix of the source (target) domain, $Y^S$ ($Y^T$) is the labels of the source (target) domain, $E=\{e_{ij}=(v_i, v_j) | v_i, v_j \in V^S\cup V^T \}$ is the beneficial-benefited edge set, where the edge $e_{ij}$ indicates $v_i$ contains useful knowledge for $v_j$ (i.e., $v_i$ is a beneficial node for $v_j$).
\end{definition}
Note that edges on the Bridged-Graph include both intra-domain edges and inter-domain edges, namely "\textbf{Knowledge Bridge}". Then we define Knowledge Bridge Learning (KBL) as follows:
\begin{definition}[Knowledge Bridge Learning]
    Knowledge Bridge Learning (KBL) is a paradigm of knowledge transfer. KBL first learns a Bridged-Graph from data to define the valid scope of knowledge transfer and then transfers knowledge on the Bridged-Graph from beneficial nodes to benefited nodes.
\end{definition}
In all three scenarios shown in Fig. \ref{fig:gkt}, KBL learns a Bridged-Graph to scope the knowledge transfer, while the difference of the learned Bridged-Graph in the three scenarios is that we will reuse the original edges in relational data which are suitable for knowledge transfer. For un-relational data scenarios, we learn Bridged-Graph from scratch with the original isolated samples. For $RD_{intra}$ scenario, we learn Bridged-Graph by adding new knowledge bridges and reusing original intra-domain edges. For $RD_{intra\&inter}$ scenario, we further reuse the original inter-domain edges.

\section{Methodology}
In this section, we introduce the motivation and architecture of our Bridged Graph Neural Network (Bridged-GNN) under the guidance of Knowledge Bridge Learning.
\subsection{Motivations of Knowledge Bridge Learning}\

\label{sec:motivation}
\subsubsection{Irregular Domain Shift in Real-world Data}\quad\\
\label{sec:motivation1}
Domain shift problem usually refers to the difference in the joint distribution (i.e., $P(X,Y)$) between the source domain and the target domain. However, directly modeling the joint distribution is difficult with few target labels, and some methods \cite{long2013jointTL, long2017deep} use pseudo labels to estimate the pseudo joint distribution, which is sensitive to noises and results in estimation bias. As one of the most representative transfer learning methods, Domain Adaptation (DA) simplifies the problem by making assumptions based on the Bayes formula:
\begin{equation}
    \nonumber
    P(X, Y)=P(X|Y)P(Y)=P(Y|X)P(X).
\end{equation}
DA methods usually assume that the source domain and target domain share invariant conditional distribution ($P(X|Y)$ or $P(Y|X)$), while having different marginal distribution ($P(X)$ or $P(Y)$)\cite{survey_da, MME, DAN}. And the mainstream framework of DA, including the unsupervised and the semi-supervised settings, is to jointly train a shared model on the source and target domains while aligning the marginal distribution discrepancy, assuming a shared posterior distribution of source and target domains during this process.

\label{sec:rand_add_edge}
\begin{figure}[h]
    \begin{minipage}[t]{0.49\linewidth}
        \centering
        \subfloat[Twitter]{\includegraphics[width=\linewidth]{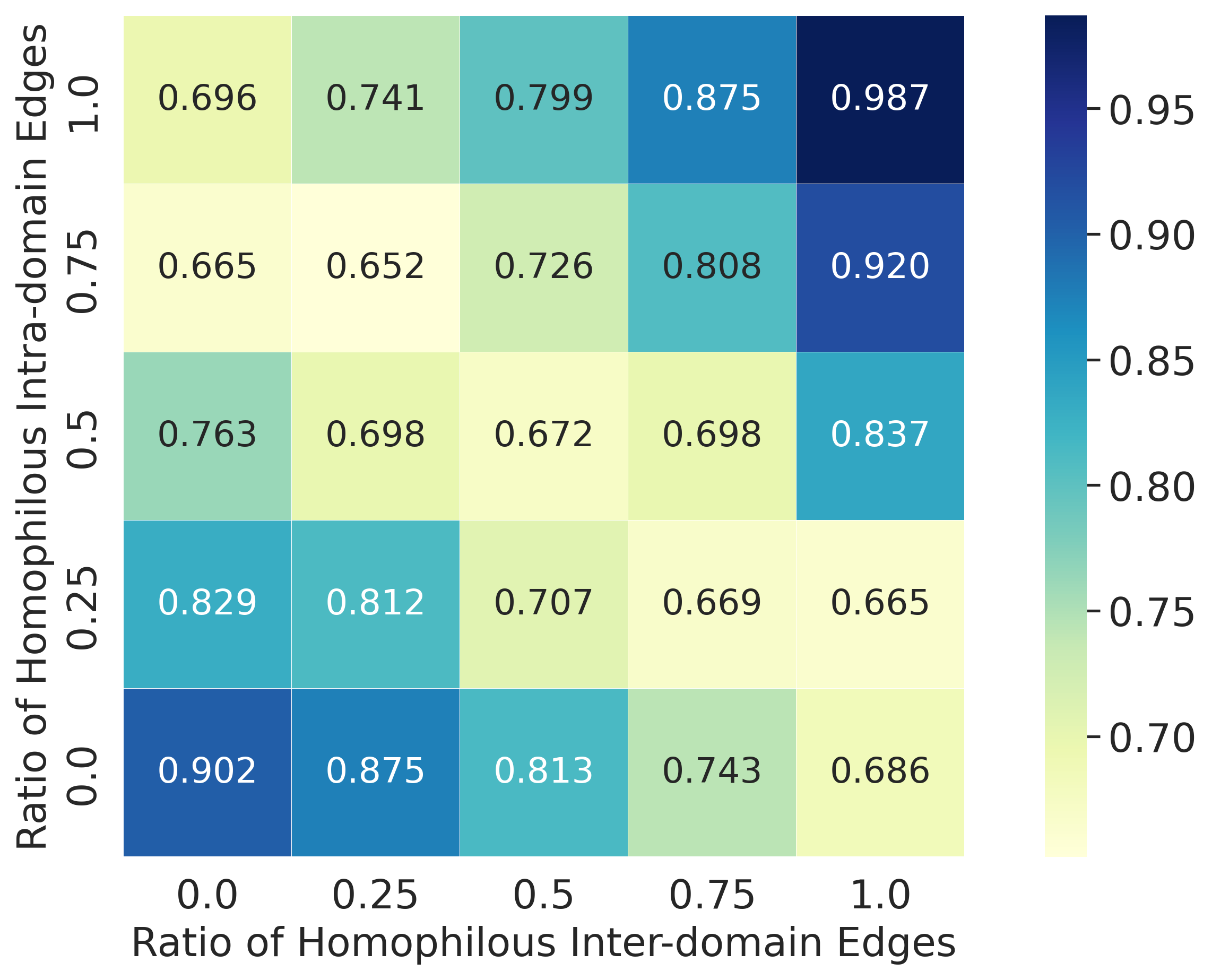}}
    \end{minipage}
    \begin{minipage}[t]{0.49\linewidth}
        \centering
        \subfloat[Office31 (A$\rightarrow$D)]{\includegraphics[width=\linewidth]{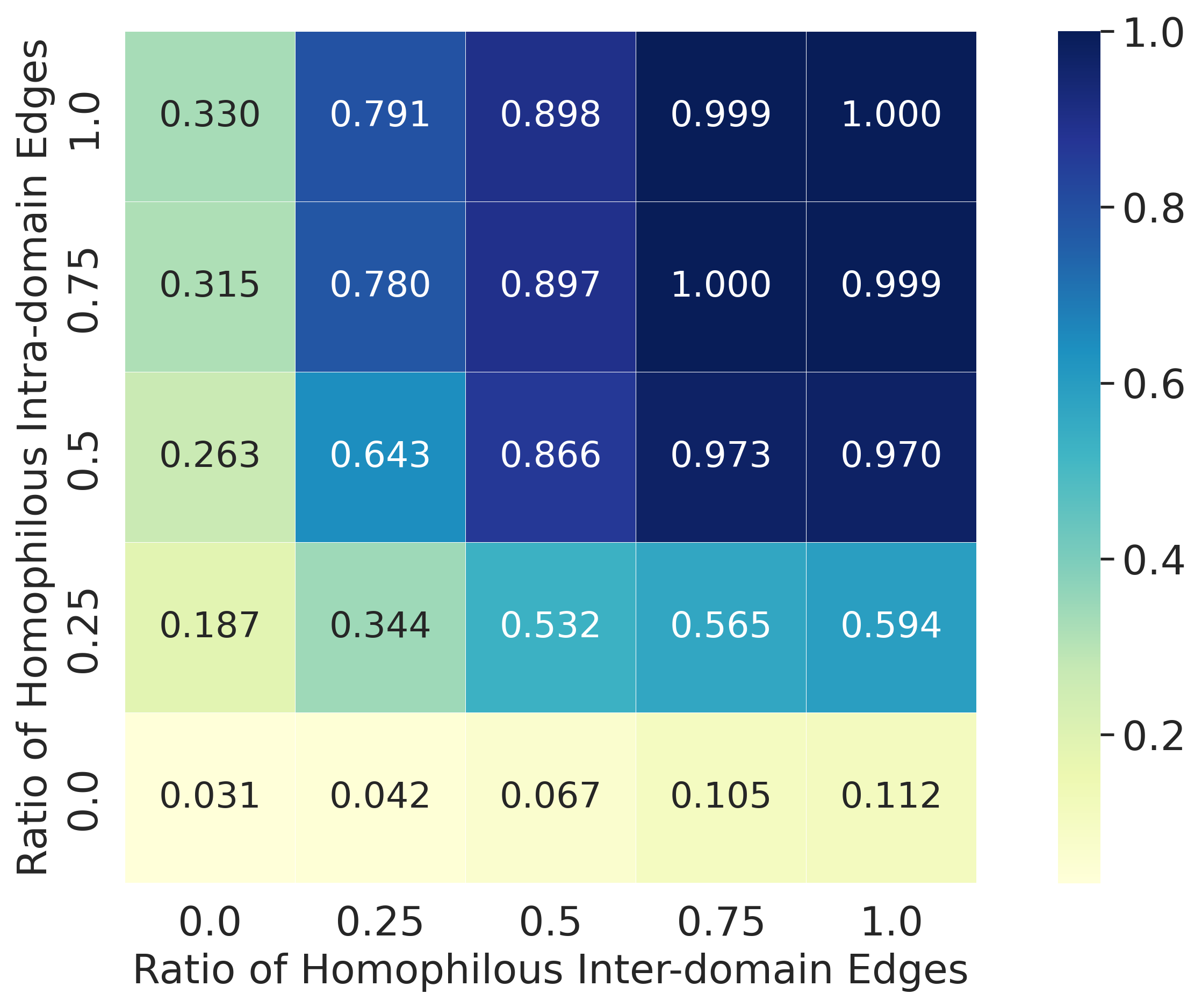}}
    \end{minipage}
    
    \caption{Results on Bridged-Graphs constructed via randomly adding edges and controlling the ratio of homophilous neighbors, including homophilous ratio of intra-domain edges (Y-axis) and inter-domain edges (X-axis).}
    \label{fig:rand_add_edge}
\end{figure}

However, we observed that this assumption is hard to be satisfied in real-world data, and the conditional distribution between the source domain and target domain may have significant differences.
 As shown in Fig. \ref{fig:domain_shift_visual}, we visualize the features of five datasets used in this paper with the T-SNE algorithm. In each scatter plot, we distinguish the samples of different classes and different domains with four different colors. For the Facebook100 dataset which has multiple classes, we only select the samples of the first two classes to present the results more clearly. Fig. \ref{fig:domain_shift_visual} (a)$\sim$(d) are from four real-world datasets, we find there exist significant domain differences on both the conditional distribution and marginal distribution in these datasets. These evidences indicate that we cannot ignore the conditional distribution differences in the real data, and motivate us to design a new paradigm. 

 As shown in Fig. \ref{fig:intro} (c), learning a shared posterior distribution ($P_{S+T}(Y|X)$) on source and target domains is sensitive to the conditional distribution shift and noise data. Different from previous transfer learning, our Knowledge Bridge Learning aims at enriching the sample information by injecting knowledge from other samples, and \textbf{changing the paradigm of learning shared posterior distribution into learning a knowledge-enhanced posterior distribution for target domain},i.e., $P_T\left(Y|X, \mathcal{K}(X)\right)$, where $\mathcal{K}(X)$ is the external knowledge of samples (source nodes on the Bridged-Graph).

\subsubsection{Effectiveness of Knowledge Transfer on Bridged-Graph}\quad\\
\label{sec:motivation2}
Knowledge Bridge Learning aims to learn a scope named Bridged-Graph and transfer knowledge via this graph. Such a graph determines the upper bound of performance improvement gained by knowledge transfer. Therefore, we first validate the effectiveness of our KBL by a synthetic Bridged-Graph. Specifically, we generated many Bridged-Graphs on two real-world datasets (Twitter \cite{xiao2020timme} and Office31 \cite{MME}) by randomly adding edges between samples while controlling the ratio of homophilous neighbors (neighbors of the same class). For each synthetic Bridged-Graph, we randomly add 8 neighbors $v_i$ as source nodes (4 nodes in $V^S$ and 4 nodes in $V^T)$ for each target domain sample $v_j\in V^T$, and only add 4 neighbors from $V^S$ as source nodes for each source domain sample $v_j\in V^S$. Then we use GraphSAGE \cite{GraphSAGE} as the Graph Knowledge Transfer module of KBL to transfer knowledge from source nodes to target nodes and record the test accuracy of node classification. As shown in Fig. \ref{fig:rand_add_edge}, when the homophilous ratio of intra-domain edges and inter-domain edges tends to 1.0 at the same time, the test accuracy nearly reaches 100\%, which demonstrates the effectiveness of our motivations (knowledge transfer on Bridged-Graph).


\begin{figure}[t]
    \centering
    \includegraphics[width=\linewidth]{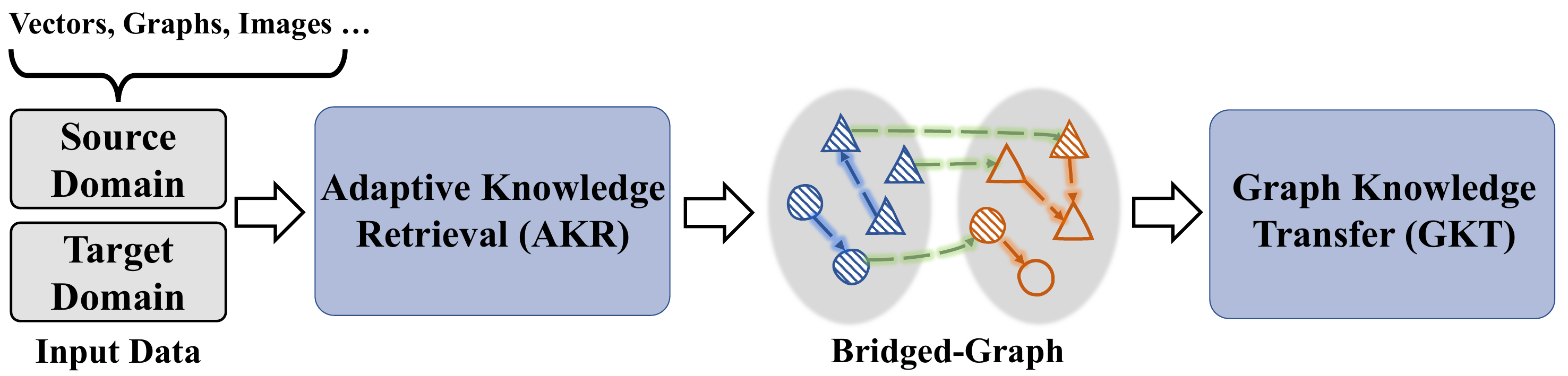}
    \captionsetup{skip=2pt}
    \caption{Overview of Bridged-GNN}
    \label{fig:bridged-gnn}
\end{figure}

\subsection{Overview of Bridged-GNN}
Guided the paradigm of Knowledge Bridge Learning in Sec.\ref{sec:def_kbl}, we propose Bridged-GNN and show its overview in Fig. \ref{fig:bridged-gnn}. Bridged-GNN is implemented in a two-stage manner which is composed of two main components: Adaptive Knowledge Retrieval (AKR) module and Graph Knowledge Transfer (GKT) module. Specifically, Bridged-GNN first learns a Bridged-Graph to determine the scope of knowledge transfer, and then uses a GNN model as plug-in to transfer knowledge on the Bridged-Graph. Bridged-GNN model can be applied to all three scenarios in Fig. \ref{fig:gkt}.

\subsection{Adaptive Knowledge Retrieval (AKR)}
In the first step, we need to learn the Bridged-Graph to determine the scope of knowledge transfer. We mainly focus on the classification tasks in this paper, and the required useful knowledge of a specified sample is mainly contained in the homophilous samples of the same class (validated in Sec. \ref{sec:rand_add_edge})\cite{bi2022make, guo2023homophily, yang2023both, bashardoust2023reducing, begga2023diffusion, pan2023beyond}. Considering the candidates of beneficial samples are from different domains, we design an Adaptive Knowledge Retrieval (AKR) module. Given arbitrarily specified sample as a query, AKR retrieves top-K beneficial samples from all candidates as intra-domain/inter-domain knowledge for the query sample.

\subsubsection{Architecture of AKR}
\begin{figure}[h]
    \centering
    \includegraphics[width=\linewidth]{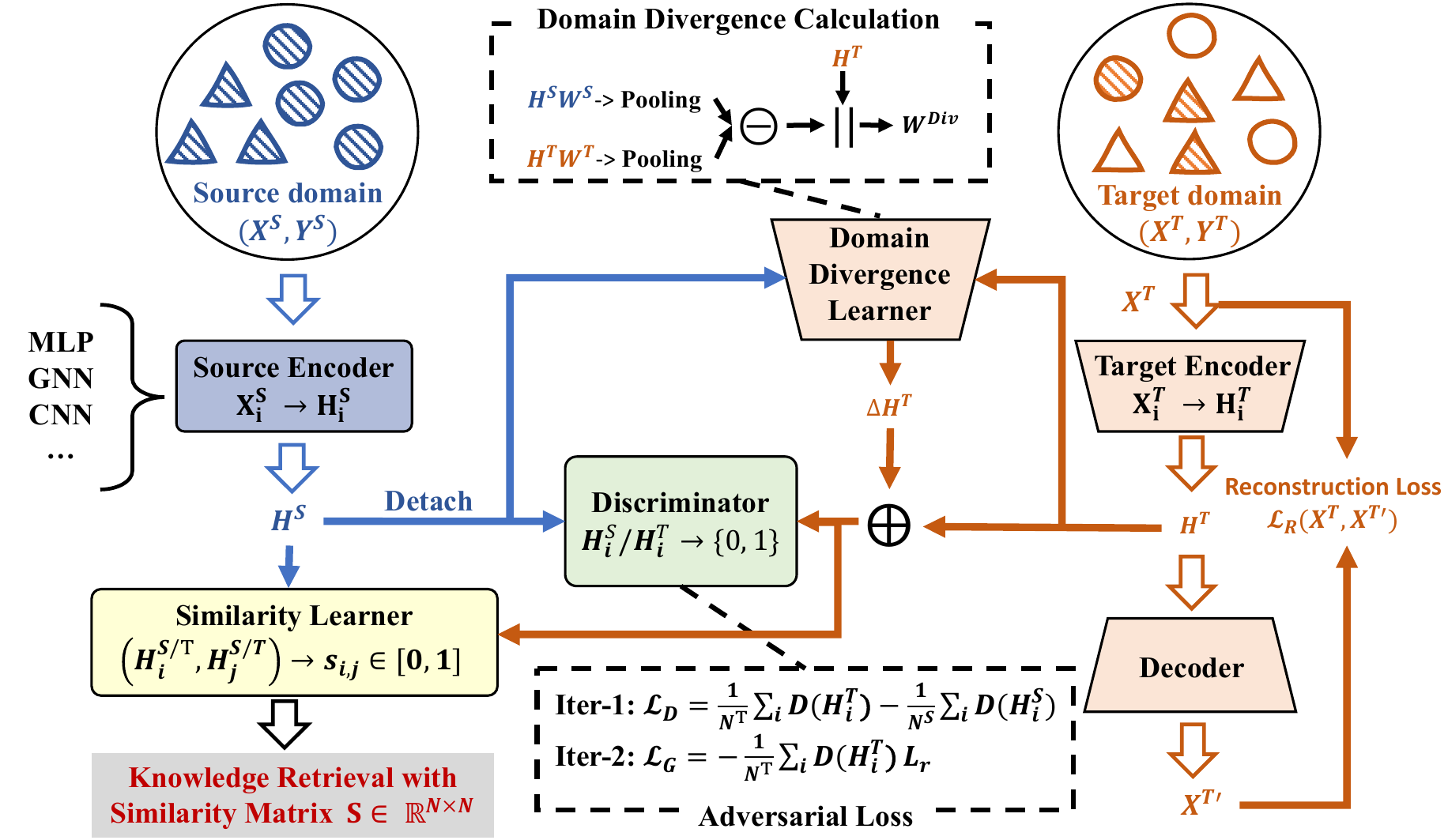}
    \captionsetup{skip=4pt}
    \caption{Architecture of the Adaptive Knowledge Retrieval (AKR) module. The blue (orange) arrows denote the data flow of the source (target) domain.}
    \label{fig:simnet}
\end{figure}

\quad\\
The AKR module first learns the similarities of samples for knowledge retrieval, which may be from the same domain (intra-domain) or different domains (inter-domain), and then retrieve beneficial samples for each sample (query) with the learned similarities to construct the Bridged-Graph. As shown in Fig. \ref{fig:simnet}, we design a twin-flow architecture for the AKR module. AKR is composed of the source encoder, the target encoder, and a discriminator with the adversarial training strategy.

First, a source encoder $\mathcal{F}^{src}(\cdot)$ and a target encoder $\mathcal{F}^{tar}(\cdot)$ are designed to encode the input data from the source domain and target domain respectively:
\begin{equation}
	  \left\{\begin{aligned}
	H^S &= \mathcal{F}^{src}(X^S, \cdots) \\
	H^T &= \mathcal{F}^{tar}(X^T, \cdots) \\
	\end{aligned}
	\right.
\end{equation}
where $X^S\in\mathbb{R}^{N^{src}\times D^{in}}$ and $X^T\in\mathbb{R}^{N^{tar}\times D^{in}}$ represent the sample features of the source domain and target domain, $H^S$ and $H^T$ denote the hidden representations of the source domain and target domain. Besides, for graph data, the input data also includes the adjacent matrix $A^{S}\in\mathbb{R}^{N^{S}\times N^{S}}$ and $A^{T}\in\mathbb{R}^{N^{T}\times N^{T}}$. According to the different types of input data (e.g., text, graph, image), we can choose appropriate backbone networks (e.g., DNN, GNN, CNN) as encoders to fit the data better.

Then we transform the representations of target domain samples ($H^T$) into the same feature space as the source domain. We achieve this by using a Domain Divergence Learner module to learn the domain difference variable (i.e., the domain difference between the source domain and target domain) which considers the personalized information of each target domain sample and the overall difference between the source domain and target domain representations:
\begin{equation}
    \Delta H^{T} = \left[ \code{Pooling}(H^SW^S) - \code{Pooling}(H^TW^T)\ || \ H^T\right]\cdot W^{Div},
\end{equation}
where \code{Pooling} is the \code{SUM} pooling function. Then we can get the transformed target domain representations denoted as $\widetilde{H}^T$:
\begin{equation}
    \widetilde{H}^T = H^T + \Delta H^{T}
\end{equation}
In order to prevent the target domain representations encoded by the target domain encoder from forgetting the original target domain information, we add a target domain decoder to form an Auto-Encoder and further optimize it by reconstruction loss:
\begin{equation}
    \mathcal{L}_R = \frac{1}{N^T} \cdot \sum_{i=1}^{N^T} \left(\code{Decoder}(X^T_i) - X^T_i\right)^2
\end{equation}

Then we use an adversarial loss to make sample pairs from intra-domain and inter-domain measured in a common semantic space. Specifically, we use a discriminator (\code{D}) to distinguish the samples from the source domain and target domain, and calculate adversarial loss  iteratively:
\begin{equation}
\label{eq:adv}
 \left\{
\begin{aligned}
    \mathcal{L}_D &= \frac{1}{N^T}\sum_{i=1}^{N^T} \code{D}(\widetilde{H_i}^T) - \frac{1}{N^S}\sum_{i=1}^{N^S} \code{D}\left(\code{detach}(H_i^S)\right) \\
    \mathcal{L}_G &= -\frac{1}{N^T}\sum_{i=1}^{N^T} \code{D}(\widetilde{H_i}^T)
\end{aligned}
\right.
\end{equation}

\begin{figure}[h]
    \centering
    \includegraphics[width=0.8\linewidth]{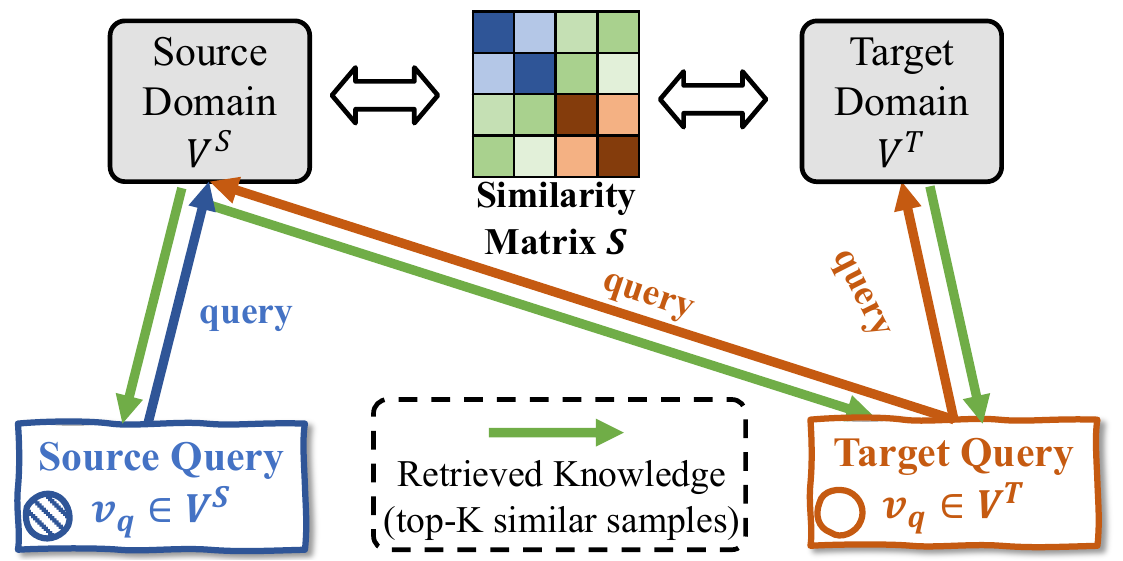}
    \caption{Knowledge retrieval (top-K beneficial samples).}
    \label{fig:retrieval}
\end{figure}

With the obtained source domain representations $H^S$ and the transformed target domain representations $\widetilde{H}^T$, which belong to the same feature space, we then train a cosine similarity  learner module to learn the pair-wise similarity of arbitrary sample pair:
\begin{equation}
    S_{i,j} = \Gamma_{\text{Cosine}}(W^\text{Sim}\widehat{H}^{S+T}_i, W^\text{Sim}\widehat{H}^{S+T}_j),\quad \widehat{H}^{S+T} = \begin{bmatrix}
        H^S \\ \widetilde{H}^T
    \end{bmatrix} 
\end{equation}
Where $S_{i,j}$ represents the similarity between nodes $v_i$ and $v_j$ ($v_i$ and $v_j$ may be from the same domain or different domains). $\Gamma_{\text{Cosine}}(\cdot, \cdot)$ represents the cosine similarity function. Finally, we get a similarity matrix  $S\in \mathbb{R}^{N\times N}$ where $N=N^S+N^T$. Then given any query sample $v_q\in \{V^S\cup V^T\}$, we can retrieve top-K (K is a hyperparameter) similar samples from $V^S\cup V^T$ as the knowledge of $v_q$:
\begin{equation}
\label{eq:knowledge}
    \mathcal{K}(v_q) = \{<v_i, v_q> | v_i\in\{V^S\cup V^T\}, S_{i,q}\in \code{top-K}(S_{:, q})\}
\end{equation}
As shown in Fig. \ref{fig:retrieval}, considering the insufficiency and low quality of target domain data, we retrieve knowledge for each source domain sample from source domain only, while retrieving knowledge for each target domain sample from both source and target domains. We optimize the AKR module by binary cross entropy (BCE) loss:

\begin{equation}
    \mathcal{L}_{CLF} = \frac{1}{N^\text{pair}}\sum_{v_i, v_j\in \{V^S\cup V^T\}} \code{BCE}\left(Y^{\text{pair}}_{ij}, \sigma (S_{ij})\right)
\end{equation}
Where $N^{\text{pair}}$ represents the number of sample pairs $<v_i, v_j>$ used for training, $ Y^{\text{pair}}_{ij}=1$ indicates that the  pair of samples $<v_i, v_j>$ belong to the same category, otherwise $Y^{\text{pair}}_{ij}=0$. $\sigma(\cdot)$ represents the \code{Sigmoid} function. Overall, we leverage $\mathcal{L}_D$ to optimize discriminator while fixing the parameters of AKR except for the discriminator, then $\mathcal{L}_{CLF}$ is leveraged together with $\mathcal{L}_{R}$ and $\mathcal{L}_{G}$  (i.e., $\mathcal{L}_{R} + \mathcal{L}_{G} + \mathcal{L}_{CLF}$) to optimize the AKR except for the discriminator. These two steps are performed iteratively.
In principle, our method can be applied to any machine learning tasks (e.g., classification and regression tasks).

\subsubsection{Scalable optimization strategy with balanced sampling}\quad\\ 
Considering that the number of possible sample pairs is the square of the universal sample size ($N^2$), which is infeasible to train with full-batch with a large $N$. Besides, sample pairs of datasets with multiple classes are extremely unbalanced, i.e., negative pairs with different labels are far more than positive pairs with the same label.

To tackle the scalability and imbalance problems, we design a scalable optimization strategy with Balanced Pair-wise  Sampling (BPS). Specifically, we use the BPS algorithm shown in Algo. \ref{algorithm:balanced_sampling} to sample $N^\text{pair}$ intra-domain sample pairs in the source domain and target domain respectively (setting $X^1=X^2=X^S, Y^1=Y^2=Y^S$ and $X^1=X^2=X^T, Y^1=Y^2=Y^T$). And we sample $N^\text{pair}$ inter-domain (source$\rightarrow$target) sample pairs (setting $X^1=X^S, X^2=X^T, Y^1=Y^S, Y^2=Y^T$). Then we get a total of $3N^\text{pair}$ pairs of samples in each iteration. We set \text{Max\_Class\_Num} to 10 for all datasets. The BPS algorithm can guarantee the same number of positive pairs and negative pairs in each mini-batch. Besides, we can set an arbitrary $N^\text{pair}$ according to the data size.
\begin{algorithm}
    \caption{Balanced Pair-wise Sampling (BPS)}
    \label{algorithm:balanced_sampling}
    \LinesNumbered
    \KwIn{$\text{Data}_1$: ($X^1$, $Y^1$),\quad $\text{Data}_2$: ($X^2$, $Y^2$),\quad
    \text{Sample\_Size}: $N^\text{pair}$,\quad \text{Max\_Class\_Num}: $M$.} 
    \KwOut{$N^\text{pair}$ pairs of sample: $\left\{<x_i\in X^1, x_j\in X^2>, Y^{\text{pair}}_{ij} =\mathbf{1}_{Y^1_i == Y^2_j} \right\}$}
    Random select $M$ categories $\mathcal{C}_y = \{y_1, y_2, \cdots y_M\}$ in all possible class of $Y^1\cup Y^2$.\\
    \code{Output} $\leftarrow \{\}$.\\
    \ForEach{$Y_\text{from}\in \mathcal{C}_y$}{
        \ForEach{$Y_\text{to}\in \mathcal{C}_y$}{
            \If{$Y_\text{from} == Y_\text{to}$}{
                $N_\text{batch} \leftarrow \lfloor\frac{0.5 * N^\text{pair}}{M}\rfloor$
            }
            \Else{
                $N_\text{batch} \leftarrow \lfloor\frac{0.5 * N^\text{pair}}{M * (M-1)}\rfloor$
            }
            1. Randomly sampling $N_\text{batch}$ samples from $\{X^1_i | Y^1_i == Y_\text{from}\}$, denoted as $\code{Batch}_1$.\\
            2. Randomly sampling $N_\text{batch}$ samples from $\{X^2_i | Y^2_i == Y_\text{to}\}$, denoted as $\code{Batch}_2$.\\
            3. $X^\text{pair}$  $\leftarrow \{ <x_i, x_j> | x_i\in \code{Batch}_1, x_j\in \code{Batch}_2 \}$\\
            \code{Output} $\leftarrow \{ \code{Output} \cup X^\text{pair} \}$.
        }
    }
\end{algorithm}

\subsection{Graph Knowledge Transfer (GKT)}
Leveraging the retrieved knowledge in the previous step, we can construct the Bridged-Graph defined in Sec. \ref{sec:def_kbl}. Then we use a GNN model to transfer knowledge on the Bridged-Graph.

\subsubsection{Conversion from Retrieved Knowledge to Bridged-Graph}\quad\\ 
\label{sec:graph_construction_strategy}
We further convert the retrieved knowledge (see Eq. \ref{eq:knowledge} and Fig. \ref{fig:retrieval}) into a Bridged-Graph. For all three scenarios of Knowledge Transfer shown in Fig. \ref{fig:gkt}, we view each sample as a node and add edges from top-K retrieved beneficial nodes to each target node, i.e. a KNN-Graph, where K is a hyperparameter set by grid search in \{4, 8, 16, 20\} in our experiments. Besides, we also reuse the original edges that are suitable for knowledge transfer in two relational data scenarios ($RD_{intra}$, $RD_{intra\&inter}$). Specifically, we remove original edges with similarities lower than the threshold $\epsilon$ (we set $\epsilon$ as the 25\% quantile number of the similarity matrix in our experiments and can adjust it according to the actual situations), and then reuse the remaining edges in the Bridged-Graph.

\subsubsection{GNN for Knowledge Transfer on Bridged-Graph.}\quad\\ 
\label{sec:selection_GNN}
 With the learned Bridged-Graph that defines the scope of knowledge transfer, we then use a Graph Neural Network model to transfer knowledge from beneficial samples to benefited samples. At this step, all message-passing-based GNNs can be used as a plug-in for the GKT module. 
 We use GNNs as our GKT module because the neighborhood aggregation framework (see Eq. \ref{eq:gnn}) of GNNs just matches our motivation (see Sec. \ref{sec:motivation1}) to learn a knowledge-enhanced posterior distribution for the target domain: 
 \begin{equation}
 \label{eq:our_motivation_with_gnn}
     P_T\left(y_i|x_i, \mathcal{K}(x_i)\right) \rightarrow P_T\left(y_i|\code{Combine}\left(x_i, \code{AGG}\left(\{x_j | v_j\in\mathcal{N}(v_i)\}\right)\right)\right)
 \end{equation}
 The \code{AGG} function  of GNN is used to aggregate the retrieved knowledge from AKR, while the \code{Combine} function of GNN is used to combine the original sample feature and the aggregated knowledge. Considering that there exist multi-domain nodes on the Bridged-Graph, we mainly use KTGNN \cite{bi2023predicting} which considers node-level domain shift as the GKT module in Bridged-GNN. And we also evaluate the performance of other mainstream GNNs (e.g., GCN, GAT, GCNII) (see Table \ref{tab:main_res_ckt} and Table \ref{tab:main_res_skt}).

\section{Experiments}
In this section, we compare Bridged-GNN with other state-of-the-art methods on the three knowledge transfer scenarios in Fig. \ref{fig:gkt}.

\subsection{Datasets}
We conduct experiments on five datasets, including four real-world datasets and  a synthetic dataset. The basic information is shown in Table \ref{tab:dataset_info}, and a brief introduction of datasets is as follows: \footnote{More detailed information of datasets used in this paper can be found at: \href{https://github.com/wendongbi/Bridged-GNN/blob/main/README.md}{https://github.com/wendongbi/Bridged-GNN/blob/main/README.md}}:

\begin{table}[h]
    \small
    \centering
    \setlength{\tabcolsep}{3.0pt}
    \caption{Basic information of the datasets used in this paper. $N^S$ and $N^T$ denote the number of samples in the source domain and target domain. $D^{in}$ is the feature dimension.}
    \label{tab:dataset_info}
    \begin{tabular}{crrrrr}
        \toprule
        Dataset & $N^S$ & $N^T$ & $D^{in}$ & Edge Num & Class Num   \\
        \midrule  
        $\text{Twitter}_{UD}$ &581 & 20230 & 300  & 0 & 2 \\
        \text{Office31} (A$\rightarrow$W) & 2817 & 888 & image & 0& 31 \\
        \text{Office31} (A$\rightarrow$D) & 2817 & 591 & image & 0& 31\\
        Sync-$UD$ & 600 & 300 & 64 & 0 & 2\\
        \midrule
        FB (Hamilton$\rightarrow$Caltech) &2314 &769 & 6 &258735  & 7\\
        FB (Howard$\rightarrow$Simmons) & 4047 & 1518 & 6 & 538924  & 7\\
        Sync-$RD_{intra}$ & 600 & 300 & 64 & 2414 & 2\\
        \midrule
        $\text{Twitter}_\text{Graph}$ &581 & 20230 & 300  & 915438 & 2 \\
        Company & 3987 & 6654 & 33  & 116785 & 2 \\
        Sync-$RD_{intra\&inter}$ & 600 & 300 & 64 & 3414 & 2\\
        \bottomrule
    \end{tabular}
\end{table}

\begin{table*}[h]
    \centering
    \renewcommand{\arraystretch}{0.85}
    \setlength{\tabcolsep}{3pt}
    \caption{Experiments of classification on target domain samples in the scenario of un-relational data ($UD$ in Fig. \ref{fig:gkt} (a)). The meaning of the model with subscripts can be found at Sec. \ref{sec:exp_setting}.} 
    \label{tab:main_res_nkt}
    \begin{tabular}{ccccccccc}
        \toprule
        Dataset &  \multicolumn{2}{c}{$\text{Twitter}_{UD}$} & \multicolumn{2}{c}{Office31 (A$\rightarrow$D)} &   \multicolumn{2}{c}{Office31 (A$\rightarrow$W)} &    \multicolumn{2}{c}{Sync-$UD$}\\
        \cmidrule(lr){2-3}\cmidrule(lr){4-5} \cmidrule(lr){6-7} \cmidrule(lr){8-9}
        Method $\backslash$ Metric & F1-macro &  AUC & F1-macro &  F1-micro & F1-macro &  F1-micro  & F1-macro &  AUC \\
        \midrule
        $\text{DNN}_\text{T}$ &  70.02\small{$\pm$1.16} &	76.68\small{$\pm$1.11}& 91.02\small{$\pm$1.20} & 91.53\small{$\pm$1.39} & 88.15\small{$\pm$0.55} &  88.68\small{$\pm$0.93}  & 81.08\small{$\pm0.92$} & 90.58\small{$\pm0.29$} \\
        $\text{DNN}_{\text{S} \rightarrow \text{T}}$ &  70.62\small{$\pm$1.03} & 76.08\small{$\pm$0.89} & 91.32\small{$\pm$0.83} & 91.58\small{$\pm$0.63} & 88.19\small{$\pm$0.63} &  88.79\small{$\pm$0.68}   & 81.56\small{$\pm1.02$} & 90.93\small{$\pm0.78$} \\
        $\text{DNN}_\text{S+T}$ &  70.85\small{$\pm$1.31} &	76.12\small{$\pm$1.38}& 90.53\small{$\pm$0.62} & 89.93\small{$\pm$0.62} & 87.83\small{$\pm$0.32} &  88.53\small{$\pm$0.59}   & 78.99\small{$\pm0.26$} & 90.06\small{$\pm1.83$} \\
        \midrule
        DANN & 74.29\small{$\pm$0.77} & 78.21\small{$\pm$1.21} & 91.68\small{$\pm$0.75}  & 92.02\small{$\pm$0.88} & 88.58\small{$\pm$0.28} &  88.92\small{$\pm$0.53}  & 81.87\small{$\pm0.98$} & 91.23\small{$\pm0.77$} \\
        DAN &  74.89\small{$\pm$0.87} & 78.08\small{$\pm$0.56} & 91.38\small{$\pm$0.52}  & 91.87\small{$\pm$0.66} & 88.69\small{$\pm$0.75} &  88.91\small{$\pm$0.98}   & 82.03\small{$\pm1.06$} & 91.61\small{$\pm0.73$} \\
        CDAN &  72.85\small{$\pm$0.96} & 76.18\small{$\pm$1.03} & 91.05\small{$\pm$0.39}  & 91.33\small{$\pm$0.42} & 88.03\small{$\pm$0.56} &  88.79\small{$\pm$0.93}   & 81.38\small{$\pm1.32$} & 90.63\small{$\pm0.91$}  \\
        MME & 77.63\small{$\pm$0.67} & 79.82\small{$\pm$1.23} & \underline{92.08\small{$\pm$0.89}}  & {92.13\small{$\pm$0.67}} & 89.08\small{$\pm$0.57} &  89.17\small{$\pm$0.73}  & 82.33\small{$\pm1.32$} & 92.16\small{$\pm0.92$}  \\
        APE &  76.21\small{$\pm$0.85} & 78.03\small{$\pm$1.38}& 91.67\small{$\pm$0.89}  & 91.88\small{$\pm$0.68} & \underline{89.23\small{$\pm$0.57}} &  \underline{89.29\small{$\pm$0.73} }  & 82.50\small{$\pm1.09$} & 92.21\small{$\pm1.22$}  \\
        $\text{S}^3\text{D}$ & \underline{78.39\small{$\pm$1.56}} & \underline{80.87\small{$\pm$1.56}}& 91.87\small{$\pm$1.09}  & \underline{92.23\small{$\pm$0.99}} & 88.98\small{$\pm$0.91} &  89.02\small{$\pm$0.92}   & \underline{83.08\small{$\pm1.58$} }& \underline{92.63\small{$\pm1.77$}} \\
        \midrule
        $\text{\textbf{Bridged-GNN}}_{\backslash\text{\textbf{KTGNN}}}$ &   $\mathbf{82.85}$\small{$\mathbf{\pm0.76}$} & $\mathbf{85.13}$\small{$\mathbf{\pm0.87}$} & \textbf{93.17}\small{$\pm$\textbf{0.39}}  & \textbf{93.33}\small{$\pm$\textbf{0.52}} & \textbf{89.67}\small{$\pm$\textbf{0.03}} & \textbf{90.31}\small{$\pm$\textbf{0.65}}   &   $\mathbf{85.47}$\small{$\mathbf{\pm0.92}$} & $\mathbf{95.16}$\small{$\mathbf{\pm1.24}$}  \\
        \bottomrule
    \end{tabular}
\end{table*}

\textbf{Twitter} \cite{bi2023predicting, xiao2020timme}: Twitter is a social network dataset that describes the social relations among politicians (source domain) and civilians (target domain) on the Twitter platform. And the goal is to predict the political tendency of civilians. By removing original edges, we also construct the $\text{Twitter}_{UD}$ dataset in un-relational data  scenarios. 

\textbf{Office31} \cite{MME}: The Office31 dataset is a mainstream image benchmark for transfer learning. There are three different domains (Amazon, Webcam, and DSLR) in this dataset. We select Webcam and DSLR, which are data-hungry, as the target domain, and use Amazon, which is rich in data, as the source domain, including \text{Office31} (A$\rightarrow$W) and \text{Office31} (A$\rightarrow$D).

\textbf{FB} \cite{facebook100}: FB (Facebook100) datasets are social networks of 100 different universities in the United States. And the goal is to predict Node identity flags. We view the social network of each university as an independent domain and select two of them to form a cross-network(domain) dataset in $RD_{intra}$ scenarios, including FB (Hamilton$\rightarrow$Caltech) and FB (Howard$\rightarrow$Simmons). 

\textbf{Company} \cite{bi2023predicting}: Company dataset is a company investment network with 10641 real-world companies in China. We regard listed companies as source domain and unlisted companies as target domain and the goal is to predict the risk status of non-listed companies. And we use this dataset in the $RD_{intra\&inter}$ scenario.

\textbf{Sync-$UD$/Sync-$RD_{intra}$/Sync-$RD_{intra\&inter}$}: We construct 3 synthetic datasets for three scenarios of GKT by randomly sampling points of source and target domains  from two distinct Multivariate Gaussian distributions, which are visualized in Fig. \ref{fig:domain_shift_visual} (e). The samples of source and target domains in this dataset are designed to have distinct conditional distribution and marginal distribution to validate the motivations described in Sec. \ref{sec:motivation}.

\subsection{Experimental Settings}
\label{sec:exp_setting}
\quad

\textbf{Data Settings:} Under the semi-supervised classification setting, we use all source domain samples  and only few target domain samples as training set for all datasets. For Twitter and Company datasets, we use the same data split as \citet{bi2023predicting}; For other datasets, we randomly select 20\% target domain samples in each class  for training, and the remaining samples are further divided into the validation set and test set in equal proportions.

\textbf{Model Settings}:
For the scenarios of $RD_{intra}$ and $RD_{intra\&inter}$, we use DNN and other GNNs, including GCN \cite{GCN}, GAT \cite{GAT}, GCNII \cite{GCN2}, OODGAT \cite{OODGAT}, KTGNN \cite {bi2023predicting},  as baselines to encode graph data. The hyperparameters settings of these models all refer to the original papers. We remove the original feature completion module of KTGNN  because the missing feature problem is not considered in this paper. For un-relational data scenarios, we also use some optimal transfer learning methods (i.e., DANN \cite{DANN}, DAN \ cite{DAN}, CDAN \cite{CDAN}, MME \cite{MME}, APE \cite{APE}, $\text{S}^3\text{D}$ \cite{S3D}), where the $\text{S}^3\text{D}$ model adopts a sample-wise distillation method for semi-supervised domain adaptation and achieves state-of-the-art performance. For transfer learning methods and the encoder of AWR in Bridged-GNN, we use ResNet34 as the backbone network for Office31 dataset and use 3-layer MLP as the backbone network for other datasets with vector features. For all transfer learning methods, we use the recommended hyperparameters in the original papers and train them all in semi-supervised learning settings. Besides, we also compare models with different training strategies:\\
    \quad(1) T:  train on the target domain only.\\
    \quad(2) S+T: train on the source and target domain concurrently.\\
    \quad(3) S$\rightarrow$T: pretrain on source domain and finetune on  target domain.\\
We use subscripts to denote a model trained under a specific strategie (e.g., $\text{DNN}_\text{T}$, $\text{DNN}_\text{S+T}$, $\text{DNN}_\text{S}\rightarrow\text{T}$) and Bridged-GNN with a specific GNN as the GKT module, e.g.,  $\text{{Bridged-GNN}}_{\backslash\text{{KTGNN}}}$.

\textbf{Evaluation Metric}: For binary-classification datasets (Twitter, Company, Synthetic datasets), we use Binary F1-Score and AUC as evaluation metrics. For multi-classification datasets (FB, Office31),  we use Macro F1-Score and Micro F1-Score as evaluation metrics. For all datasets, we evaluate the models by their performance of classifying test samples in the target domain.

\begin{table*}[h]
    \centering
    \caption{Experiments of node classification on target domain samples in the scenarios of relational data with only intra-domain relations ($RD_{intra}$ in Fig. \ref{fig:gkt} (b)).  The meaning of the model with subscripts can be found at Sec. \ref{sec:exp_setting}.
    }
    \label{tab:main_res_ckt}
    \begin{tabular}{ccccccc}
        \toprule
        Dataset &  \multicolumn{2}{c}{FB (Hamilton$\rightarrow$Caltech)} & \multicolumn{2}{c}{FB (Howard$\rightarrow$Simmons)} & \multicolumn{2}{c}{Sync-$RD_{intra}$}\\
        \cmidrule(lr){2-3}\cmidrule(lr){4-5}\cmidrule(lr){6-7}
       Method    & F1-Macro &  F1-Micro  & F1-Macro  &  F1-Micro & F1-macro  &  AUC \\
        \midrule
        $\text{GCN}_{\text{T}}$
        &76.25\small{$\pm$0.35} & 85.38\small{$\pm$0.17} & 55.02\small{$\pm0.49$} & 89.55\small{$\pm0.14$} & 89.63\small{$\pm0.77$} & 94.36\small{$\pm0.14$}\\
        $\text{GCN}_{\text{S+T}}$ & 75.11\small{$\pm$2.57} & 85.50\small{$\pm$1.02} & 50.35\small{$\pm0.57$} & 89.49\small{$\pm0.21$} & 86.66\small{$\pm0.90$} & 93.88\small{$\pm0.30$}\\
        $\text{Bridged-GNN}_{\backslash\text{GCN}}$ & \textbf{87.62}\small{$\pm$\textbf{1.78}} & \textbf{90.78}\small{$\pm$\textbf{1.14}} & \textbf{62.03}\small{$\pm\mathbf{2.56}$} & \textbf{90.29}\small{$\pm\mathbf{0.52}$} & \textbf{93.88}\small{$\pm$ \textbf{0.77}} & \textbf{96.80}\small{$\pm$\textbf{0.33}}\\
        \midrule
        $\text{GAT}_{\text{T}}$ & 75.13\small{$\pm$0.65} & 84.78\small{$\pm$0.53} & 55.31\small{$\pm0.52$} & 89.78\small{$\pm0.72$} & 90.02\small{$\pm1.09$} & 94.55\small{$\pm0.14$}\\
        $\text{GAT}_{\text{S+T}}$ & 74.23\small{$\pm$1.87} & 84.31\small{$\pm$1.31} & 51.89\small{$\pm0.62$} & 89.98\small{$\pm0.85$} & 87.53\small{$\pm0.70$} & 93.63\small{$\pm1.21$}\\
        $\text{Bridged-GNN}_{\backslash\text{GAT}}$ & \textbf{86.32}\small{$\pm$\textbf{1.32}} & \textbf{90.58}\small{$\pm$\textbf{1.08}} & \textbf{60.67}\small{$\pm$\textbf{1.58}} & \textbf{90.57}\small{$\pm$\textbf{0.17}} & \textbf{94.67}\small{$\pm$ \textbf{0.58}} & \textbf{97.97}\small{$\pm$\textbf{0.76}}\\
        \midrule
        $\text{GCNII}_{\text{T}}$ & 77.23\small{$\pm$1.03} & 86.89\small{$\pm$0.68} & 58.19\small{$\pm0.77$} & 90.08\small{$\pm0.23$} & 90.67\small{$\pm0.85$} & 94.21\small{$\pm0.63$}\\
        $\text{GCNII}_{\text{S+T}}$ & 75.83\small{$\pm$1.03} & 85.38\small{$\pm$1.12} & 55.89\small{$\pm0.98$} & 90.33\small{$\pm0.87$}  & 87.61\small{$\pm1.01$} & 94.32\small{$\pm1.21$}\\
        $\text{Bridged-GNN}_{\backslash\text{GCNII}}$ & \textbf{86.01}\small{$\pm$\textbf{0.93}} & \textbf{90.03}\small{$\pm$\textbf{1.33}} & \textbf{62.87}\small{$\pm$\textbf{1.55}} & \textbf{90.67}\small{$\pm$\textbf{0.53}} & \textbf{94.32}\small{$\pm$ \textbf{0.58}} & \textbf{97.78}\small{$\pm$\textbf{0.62}}\\
        \midrule

        $\text{Bridged-GNN}_{\backslash\text{KTGNN}}$ &  $\textbf{88.23}$\small{$\pm$\textbf{0.88}} & 
        $\textbf{91.51}$\small{$\pm$\textbf{0.79}}  & 
        $\textbf{66.37}$\small{$\pm$\textbf{0.23}} & 
        $\textbf{91.10}$\small{$\pm$\textbf{0.73}}  & 
        $\textbf{95.55}$\small{$\pm$\textbf{0.49}} & 
        $\textbf{97.97}$\small{$\pm$\textbf{0.33}} \\
        \bottomrule
    \end{tabular}
\end{table*}

\begin{table*}[h]
    \centering
    \caption{Experiments of node classification on target domain samples in the scenarios of relational data with both intra-domain and inter-domain relations ($RD_{intra\&inter}$ in Fig. \ref{fig:gkt} (c)).  The meaning of the model with subscripts can be found at Sec. \ref{sec:exp_setting}.} 
    \label{tab:main_res_skt}
    \begin{tabular}{ccccccc}
        \toprule
        Dataset & \multicolumn{2}{c}{$\text{Twitter}_\text{Graph}$} & \multicolumn{2}{c}{Company} & \multicolumn{2}{c}{Sync-$RD_{intra\&inter}$}\\
        \cmidrule(lr){2-3}\cmidrule(lr){4-5}\cmidrule(lr){6-7}
        Method $\backslash$ Metric  & F1-macro &  AUC  & F1-macro  &  AUC  & F1-macro  &  AUC  \\
        \midrule
        $\text{DNN}_\text{T}$ & 70.02\small{$\pm$1.16} &	76.68\small{$\pm$1.11} & 57.01\small{$\pm0.42$} & 56.35\small{$\pm0.55$} & 81.08\small{$\pm0.92$} & 90.58\small{$\pm0.29$}\\
        $\text{DNN}_{\text{S}\rightarrow\text{T}}$ & 70.62\small{$\pm$1.03} & 76.08\small{$\pm$0.89} & 57.63\small{$\pm0.42$} & 57.01\small{$\pm0.55$} & 81.56\small{$\pm1.02$} & 90.93\small{$\pm0.78$}\\
        $\text{DNN}_\text{S+T}$ & 70.85\small{$\pm$1.31} &	80.12\small{$\pm$1.38} & 57.23\small{$\pm0.38$} & 56.71\small{$\pm0.43$} & 78.99\small{$\pm0.26$} & 91.06\small{$\pm1.83$}\\
        GCN & 80.19\small{$\pm$0.87} & 86.88\small{$\pm$1.23} & 57.60\small{$\pm0.49$} & 57.83\small{$\pm0.93$}  & 79.67\small{$\pm2.66$} & 90.53\small{$\pm0.59$}\\
        GAT & 82.31\small{$\pm$1.56} & 87.88\small{$\pm$1.21} & 57.98\small{$\pm$0.43} & 58.40\small{$\pm$0.43}  & 82.78\small{$\pm1.26$} & 93.08\small{$\pm2.03$}\\
        GraphSAGE& 85.35\small{$\pm$1.32} &	92.07\small{$\pm$1.08} & 59.21\small{$\pm0.93$} & 61.55\small{$\pm1.78$}  & 91.69\small{$\pm1.78$} & 95.87\small{$\pm0.63$}\\
        GCNII& 83.85\small{$\pm$1.17} &	90.67\small{$\pm$1.32} & 58.21\small{$\pm0.88$} & 60.06\small{$\pm0.63$}  & 89.98\small{$\pm0.26$} & 95.07\small{$\pm0.31$}\\
        OODGAT & 85.42\small{$\pm$2.01} &	92.67\small{$\pm$1.60} & 60.05\small{$\pm0.89$} & 61.37\small{$\pm0.92$}  & 88.79\small{$\pm1.77$} & 93.58\small{$\pm1.33$}\\
        KTGNN & \underline{86.60\small{$\pm$1.21}} &	\underline{93.38\small{$\pm$1.60}} & \underline{61.78\small{$\pm0.53$}} & \underline{63.55\small{$\pm0.65$}} &
        \underline{93.33\small{$\pm$0.37}} &	\underline{96.38\small{$\pm$0.28}}\\
        \midrule
        $\text{\textbf{Bridged-GNN}}_{\backslash\text{\textbf{KTGNN}}}$ & 
        $\textbf{87.72}$\small{$\mathbf{\pm0.80}$} & $\textbf{95.32}$\small{$\mathbf{\pm1.03}$}  & $\textbf{63.06}$\small{$\mathbf{\pm0.49}$} & $\textbf{65.61}$\small{$\mathbf{\pm0.58}$}  & 
        $\textbf{96.67}$\small{$\mathbf{\pm0.13}$} &
        $\textbf{99.56}$\small{$\mathbf{\pm0.37}$}\\
        \bottomrule
    \end{tabular}
\end{table*}

\subsection{Main Experiments}
Analysis of experiments on the three main scenarios is as follows.

\subsubsection{Results of Knowledge Transfer in $UD$ scenarios}\quad\\
As shown in Table \ref{tab:main_res_nkt}, we conduct experiments on four un-relational datasets, including three real-world datasets ($\text{Twitter}_{UD}$, Office31 (A$\rightarrow$D),   Office31 (A$\rightarrow$W)) and one synthetic dataset  Sync-$UD$. Our $\text{Bridged-GNN}_\text{KTGNN}$ gains significant improvements of classification performance on all datasets, e.g., gains $4.46\%$ improvements of F1-macro on $\text{Twitter}_{UD}$ dataset. Compared with other baseline methods and state-of-the-art transfer learning methods, our method gains the best performance in the scenarios of knowledge transfer on un-relational data ($UD$).

\subsubsection{Results of Knowledge Transfer in $RD_{intra}$ scenarios}\quad\\
As shown in Table \ref{tab:main_res_ckt}, we conduct experiments on three cross-network datasets, including two real-world datasets from Facebook social networks (FB (Hamilton$\rightarrow$Caltech),   FB (Howard$\rightarrow$Simmons)) and one synthetic dataset  Sync-$RD_{intra}$. Considering most graph neural networks are not designed for cross-network graph representation learning, we compare our Bridged-GNN with two other learning frameworks (T and S+T, see Sec. \ref{sec:exp_setting}) to validate the gain of performance caused by Bridged-GNN, i.e., our idea of "{bridging cross-networks samples}". The results show that all variants of $\text{Bridged-GNN}_*$ combined with a specifical GNN model (e.g., GCN, GAT, GCNII, KTGNN) gain significant improvements in classification performance on all datasets. By bridging originally independent graphs, our method gains the best performance in scenarios of intra-domain relational data ($RD_{intra}$).

\subsubsection{Results of Knowledge Transfer in $RD_{intra\&inter}$ scenarios}\quad\\
As shown in Table \ref{tab:main_res_skt}, we conduct experiments on two real-world graph datasets ($\text{Twitter}_\text{Graph}$, Company) and a synthetic dataset  Sync-$RD_{intra\&inter}$. Compared with other mainstream GNNs, our $\text{Bridged-GNN}_\text{KTGNN}$ gains significant improvements in classification tasks on all graph datasets. By building a Bridged-Graph based on the original graph structure, our method gains the best performance in the scenarios of relational data with both intra-domain and inter-domain relations ($RD_{intra\&inter})$.

\begin{figure}[h]
    \begin{minipage}[t]{0.49\linewidth}
        \centering
        \subfloat[$\text{Twitter}_{UD}$]{\includegraphics[width=\linewidth]{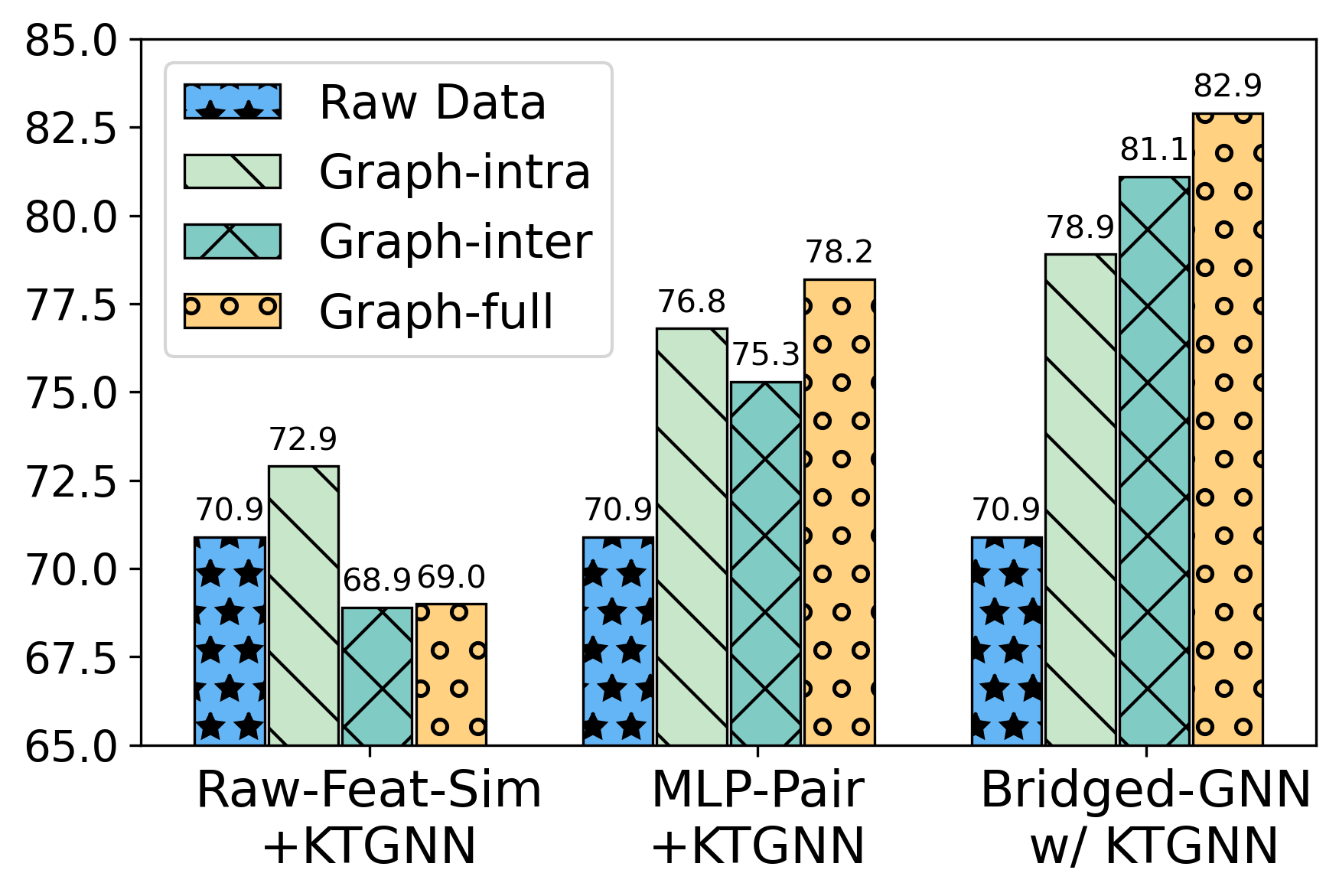}}
    \end{minipage}
    \begin{minipage}[t]{0.49\linewidth}
        \centering
        \subfloat[Office31 (A$\rightarrow$D)]{\includegraphics[width=\linewidth]{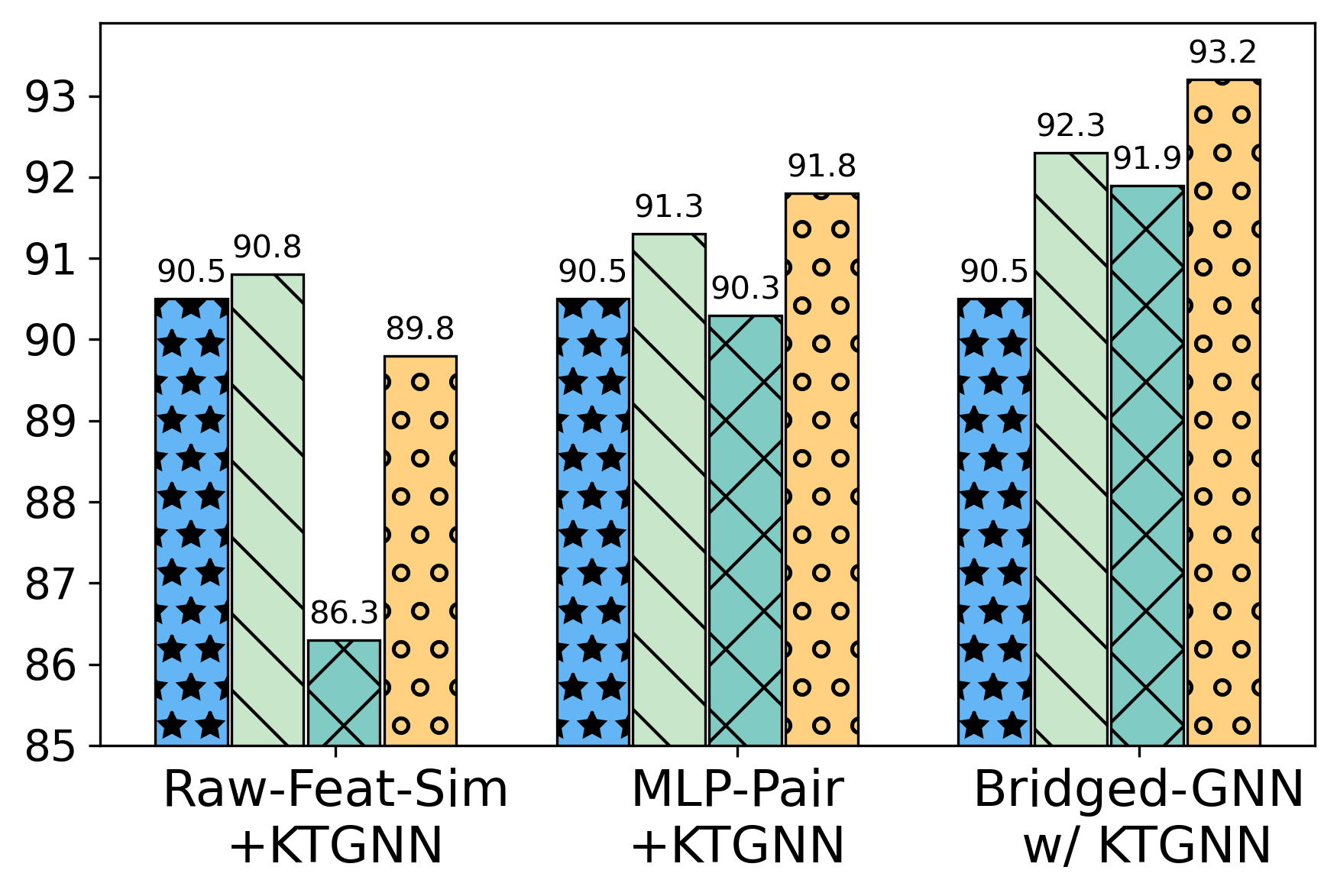}}
    \end{minipage}
    
    \caption{Ablation study of Bridged-Graphs with different components, including (1) Raw Data (un-relational) without  edges, (2) Graph-intra with only intra-domain edges, (3) Graph-inter with only inter-domain edges, (4) Graph-full with both intra-domain and inter-domain edges. Y-axis denotes the performance of KTGNN.}
    \label{fig:ablation_graph}
\end{figure}

\begin{table}[h]
    \centering
    \small
    \setlength{\tabcolsep}{2.0pt}
    \renewcommand{\arraystretch}{0.85}
    \caption{Ablation study by replacing AKR module in $\text{Bridged-GNN}_\text{KTGNN}$ with other similarity learners. We evaluate their performance of intra-domain/inter-domain sample-pair classification and target node classification on Bridged-Graphs (denoted by "NC on Bridged-Graph").}
    \label{tab:ablation_simnet}
    \begin{tabular}{cccc}
        \toprule
        Replace of AKR & F1-macro@  & $\text{Twitter}_{UD}$ & Office31 (A$\rightarrow$D)\\
        \midrule
        \multirow{3}*{\makecell[c]{Raw Feature \\ Similarity}} & Intra-domain Pair & $47.40$\small{$\pm0.61$} & $76.72$\small{$\pm0.24$} \\
        ~& Inter-domain Pair & 38.28\small{$\pm0.11$} & 60.45\small{$\pm0.23$}	 \\
        ~& NC on Bridged-Graph & 69.01\small{$\pm0.56$} & 89.83\small{$\pm0.55$}	 \\
        \midrule
        \multirow{3} *{\makecell[c]{Point-wise \\
        DNN Classifier}} & Intra-domain Pair  & 65.49\small{$\pm1.83$} & 87.56\small{$\pm0.76$} \\
        ~& Inter-domain Pair & 60.38\small{$\pm1.21$} & 81.32\small{$\pm0.59$}	 \\
        ~& NC on Bridged-Graph & 71.60\small{$\pm0.87$} & 91.03\small{$\pm0.93$}	 \\
        \midrule
        \multirow{3} *{\makecell[c]{Pair-wise \\
        DNN Classifier}} & Intra-domain Pair  & \underline{72.65\small{$\pm0.98$}} & \underline{92.89\small{$\pm0.55$}} \\
        ~& Inter-domain Pair & \underline{70.07\small{$\pm1.09$}} & \underline{92.66\small{$\pm0.67$}}	 \\
        ~& NC on Bridged-Graph & \underline{78.15\small{$\pm0.68$}} & \underline{91.89\small{$\pm0.53$}}	 \\
        \midrule
        \multirow{3} *{\makecell[c]{AKR of \\
        Bridged-GNN}} & Intra-domain Pair &$\mathbf{80.76}$\small{$\mathbf{\pm1.63}$} &  $\mathbf{96.09}$\small{$\mathbf{\pm0.72}$} \\
        ~& Inter-domain Pair &$\bm{78.13}$\small{$\bm{\pm1.28}$} & $\bm{95.41}$\small{$\bm{\pm 0.71}$}  \\
        ~& NC on Bridged-Graph & $\mathbf{82.85}$\small{$\mathbf{\pm0.76}$} & $\bm{93.17}$\small{$\bm{\pm 0.39}$}  \\
        \bottomrule
    \end{tabular}
\end{table}

\subsection{Ablation Study}
We dive into the mechanisms of Bridged-GNN by ablation studies.
\subsubsection{Effects of Adaptive Knowledge Retrieval (AKR) module}\quad\\
\label{sec:ablation_akr}
We conduct ablation studies on the AKR module of Bridged-GNN by comparing our AKR module with other traditional methods of learning sample-pair similarity \cite{graph_adaptive_knowledge_transfer, chen2021weak, wu2023curvagn}, and we evaluate their F1-score of classifying whether pairs of samples belong to the same class or not.  As shown in Table \ref{tab:ablation_simnet}, Raw Feature Similarity denotes calculating the pair-wise similarity with the raw features. Point-wise DNN Classifier means we directly train a classifier to classify each sample and then calculate the pair-wise similarity with the sample logits (output of the last layer). Pair-wise DNN Classifier means we directly train a DNN-based binary classifier to classify a pair of samples whether they belong to the same class. As shown in Table \ref{tab:ablation_simnet},  our AKR module can always gain the best performance on pair-wise classification and node classification on Bridged-Graph.

\subsubsection{Effects of Intra-domain/Inter-domain Connections}\quad\\
As shown in Fig. \ref{fig:ablation_graph}, we also validate the effects of intra-domain/inter-domain connections when constructing Bridged-Graphs with the retrieved knowledge (top-K beneficial samples). 
The results show that \textbf{the inter-domain connections establishes the bridge for knowledge transfer from the source domain to the target domain, while the intra-domain connections further enlarge the scope and effects of knowledge transfer}.

\subsection{Hyperparameter Sensitivity Analysis}
\begin{figure}[h]
    \begin{minipage}[t]{0.49\linewidth}
        \centering
        \subfloat[$\text{Twitter}_{UD}$]{\includegraphics[width=\linewidth]{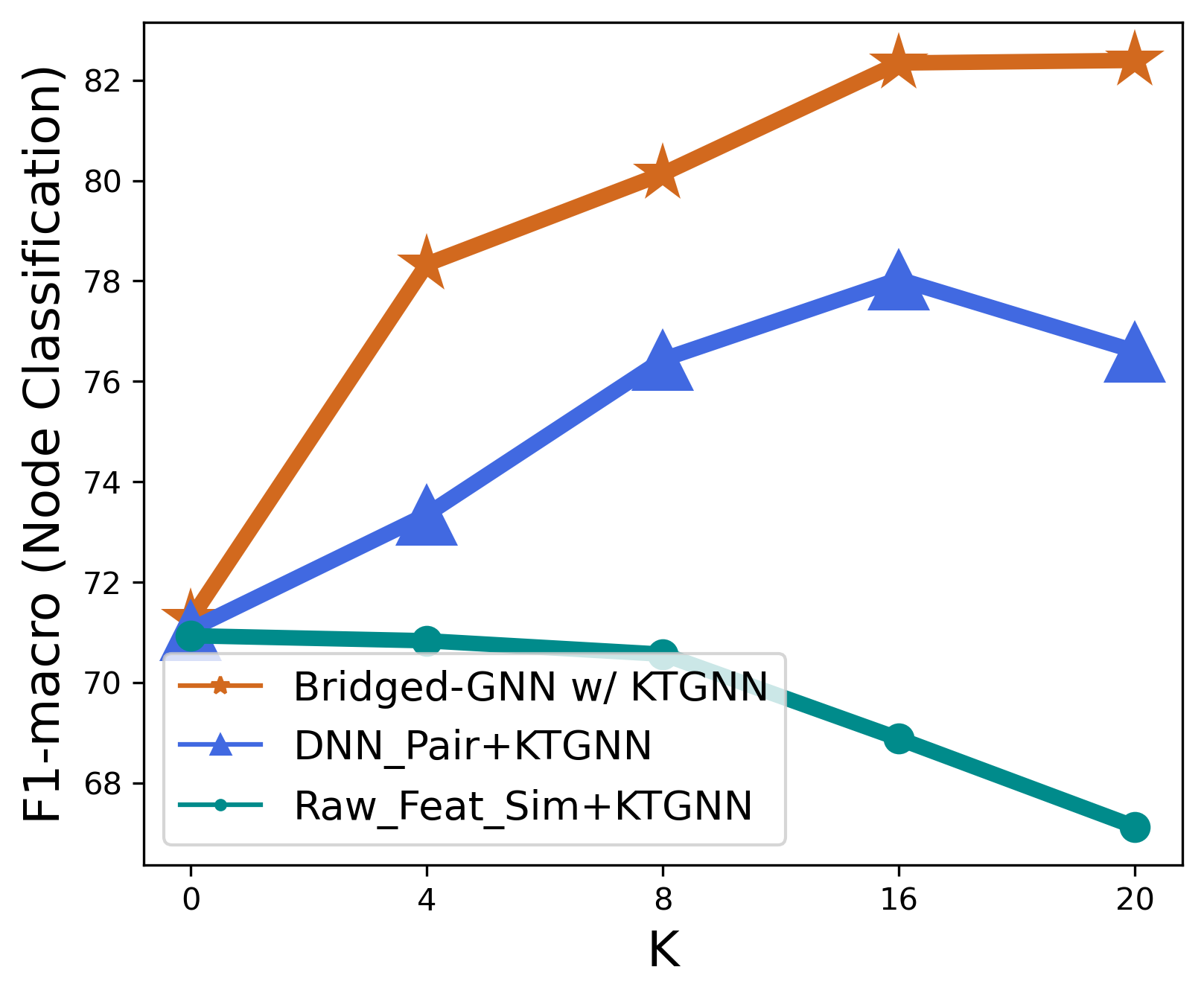}}
    \end{minipage}
    \begin{minipage}[t]{0.49\linewidth}
        \centering
        \subfloat[Office31 (A$\rightarrow$D)]{\includegraphics[width=\linewidth]{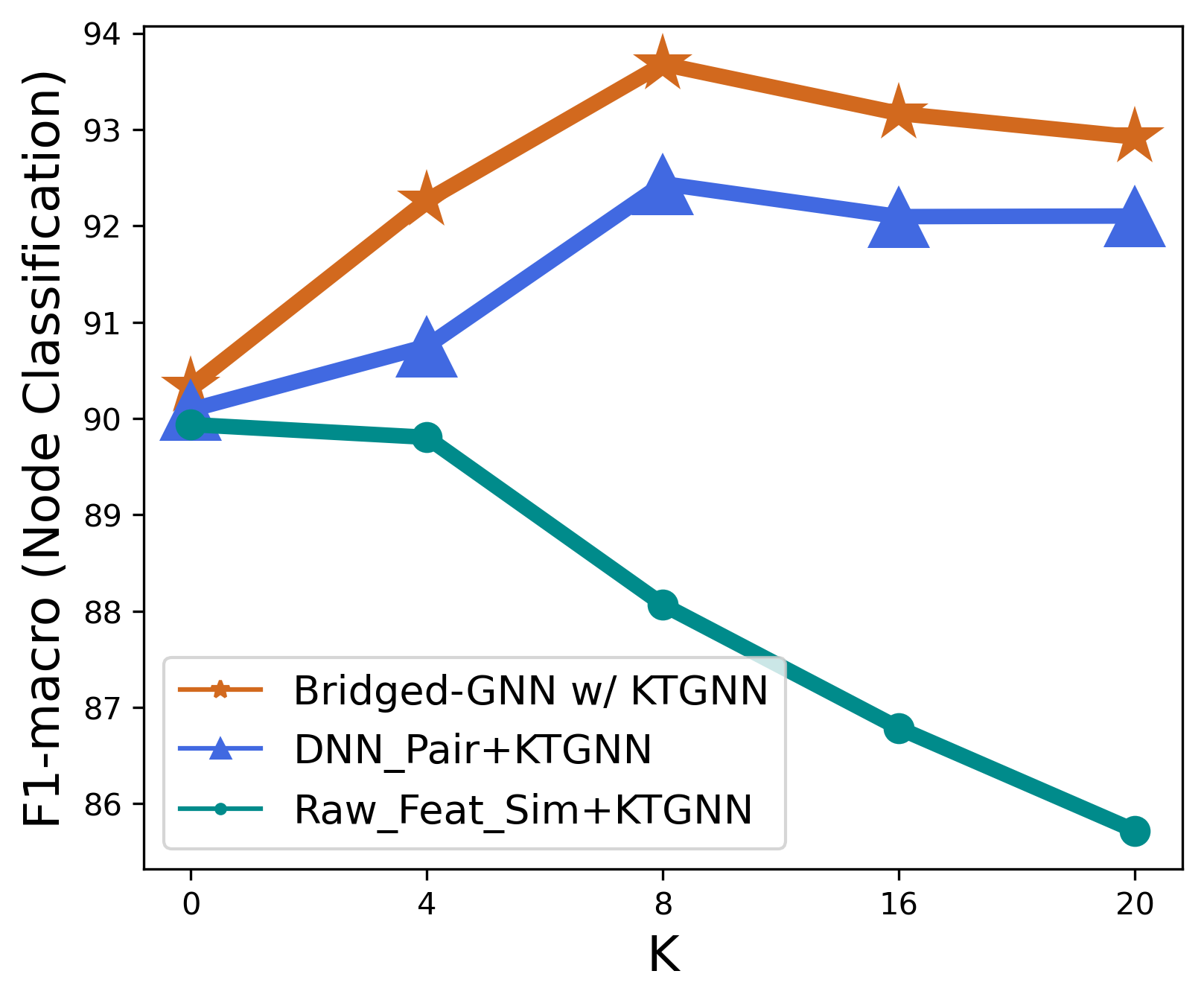}}
    \end{minipage}
    
    \caption{Analysis on hyperparameter \code{K}, which denotes the number of retrieved knowledgeable nodes for each target node and controls the density of the Bridged-Graph.}
    \label{fig:exp_hyper_k}
\end{figure}

We analyze the effects of an important hyperparameter K which controls the edge density of the Bridged-Graph. K achieves the trade-off between the quantity and quality of knowledge transfer, i.e., a larger K leads to a larger quantity of knowledge transfer, while the quality of the transferred knowledge is lower. Fig. \ref{fig:exp_hyper_k} shows the final classification results of Bridged-GNN with different values of K (X-axis in Fig. \ref{fig:exp_hyper_k}). By  replacing the AKR module in Bridged-GNN with other similarity learners (see Sec. \ref{sec:ablation_akr}), we observe that the full Bridged-GNN model can always achieve the best performance, and the results of our model are more robust as K increases.

\section{Related Works}
Transfer learning is the mainstream framework of implementing knowledge transfer to alleviate data-hungry. Domain Adaptation (DA), as the representative transfer learning branch, best matches the topic of knowledge transfer in this paper and can be broadly divided into methods with shallow and deep architectures. Shallow DA methods \cite{shallowDA_1, shallowDA_2}  mainly utilize feature-based or instance-reweight-based strategies to minimize the inter-domain distribution distance measured by some metrics (e.g., MMD \cite{MMD}, CORAL \cite{CORAL}). Deep DA methods \cite{hu2023inclusive, cao2022contrastive, kang2023node, hu2023intelligent, howard2022cross, su2022cross, yue2022contrastive, jin2021deep} use neural networks to learn shared or separated encoders for both source and target domains, while eliminating the domain gap. However, these methods usually learn a shared posterior distribution between domains and require a relatively closer inter-domain divergence, thus having severe limitations. Traditional DA methods usually refer to unsupervised DA (UDA), while recent studies \cite{feng2021cann,MME, survey_da, S3D} prove that UDA has significant defects in substantial domain-shift scenarios and propose to gain better performance via semi-supervised DA.

Graph Neural Networks (GNNs) have gained powerful performance on modeling graph data. Existing mainstream GNNs \cite{xu2021towards, xu2023simple, bi2023mm, chen2023multi, lin2014large, GWNN, lin2020initialization, wsdm2022guo, gopalakrishnan2023graph} follow the message-passing framework, where node representations are updated based on features aggregated from neighbors.  However, most of the existing GNNs are built on the IID hypothesis, i.e., all samples belong to the same data distribution. And the IID hypothesis is hard to be satisfied in many real scenarios, making the model performance degrade significantly. Recently, some GNNs for OOD scenarios \cite{zhao2023graphglow, sadeghi2021distributionally, kang2022neural, huang2019dot, zhang2023company, zhang2023company, huangend} have been designed, and studies \cite{liu2021learning,bi2023predicting, guo2023manipulating, wu2022towards} prove that GNNs can complete inter-domain knowledge transfer on graphs well. However, they strictly rely on high-quality graph structures and cannot be applied to non-graph ($UD$) and cross-graph ($RD_{intra}$) scenarios.







\section{Conclusion}
In this paper, we redefine the paradigm of knowledge transfer by Knowledge Bridge Learning (KBL) to solve the data-hungry problem. Compared with the existing transfer learning paradigm with strong assumptions, KBL learns a knowledge-enhanced posterior distribution of the target domain by defining the scope of knowledge transfer first and then transferring knowledge with GNNs. Correspondingly, we propose a novel Bridged-GNN model under the paradigm of KBL to conduct sample-wise knowledge transfer in both un-relational and relational data-hungry scenarios. Comprehensive experiments demonstrate that Bridged-GNN under KBL paradigm outperforms existing methods by a large margin.

\begin{acks}
    This work was supported by the National Natural Science Foundation of China (Grant No.U21B2046, No.62202448, No.61902380, No.61802370,  and No.U1911401), the Beijing Nova Program (No. Z201100006820061) and China Postdoctoral Science Foundation (2022M713208).
\end{acks}


\newpage
\bibliographystyle{ACM-Reference-Format}
\bibliography{sample-base}




\end{document}